\documentclass[10pt,twocolumn,letterpaper]{article}

\usepackage[pagenumbers]{paper} %

\usepackage{algorithm}
\usepackage{algpseudocode}

\newcommand{\kmin}{k_{\min}}
\newcommand{\tdup}{\tau_{\mathrm{dup}}}
\newcommand{\tsplit}{\tau_{\mathrm{split}}}
\newcommand{\Singh}{Singh \etal\ (2012)~\cite{singh2012discriminative}}
\newcommand{\Rooy}{Rooy (2019)~\cite{rooy2019posters}}
\newcommand{\myparagraph}[2][.]{\par\vspace{1pt}\noindent\textbf{#2#1}\hspace{0.5em}}   %
\newcommand{\mdot}[1]{\textcolor[HTML]{#1}{\raisebox{-0.2ex}{\scalebox{1.7}{\textbullet}}}}
\newcommand{\dotKmeans}{\mdot{0072B2}}\newcommand{\dotSVM}{\mdot{D55E00}}\newcommand{\dotTSNE}{\mdot{009E73}}
\newcommand{\dotDBSCAN}{\mdot{000000}}\newcommand{\dotVLM}{\mdot{CC79A7}}

\newcommand{\TODO}[1]{\textbf{\color{red}[TODO: #1]}}

\usepackage{microtype}   %

\newif\ifpending \pendingtrue
\newsavebox\pendbox
\newcommand{\pendingfig}[3][width=\linewidth]{%
  \IfFileExists{#2}{\sbox\pendbox{\includegraphics[#1]{#2}}}{\sbox\pendbox{\rule{0pt}{5cm}\rule{\linewidth}{0pt}}}%
  \ifpending\fbox{\parbox[c][\dimexpr\ht\pendbox+\dp\pendbox-2\fboxsep-2\fboxrule][c]{\dimexpr\wd\pendbox-2\fboxsep-2\fboxrule}{\centering
    \color{gray}\textbf{Figure placeholder: #3}\\[2pt]\small not accepted yet; edit it on the figure site}}%
  \else\usebox\pendbox\fi}

\usepackage{placeins}

\providecommand{\crfill}[1]{}
\providecommand{\crplace}[1]{}

\makeatletter
\g@addto@macro\normalsize{\setlength{\abovedisplayskip}{4pt plus 1pt minus 2pt}\setlength{\belowdisplayskip}{4pt plus 1pt minus 2pt}%
  \setlength{\abovedisplayshortskip}{2pt plus 1pt}\setlength{\belowdisplayshortskip}{3pt plus 1pt minus 1pt}}
\makeatother
\AtBeginDocument{\crefformat{equation}{eq.~#2#1#3}\crefmultiformat{equation}{eqs.~#2#1#3}{ and~#2#1#3}{, #2#1#3}{ and~#2#1#3}}

\usepackage{afterpage}

\usepackage{xspace}   %

\newcommand{\CineName}{CINEMATERIAL-161K\xspace}

\newcommand{\NKept}{160{,}742}

\newcommand{\WinPerPoster}{13}
\newcommand{\YearMin}{1888}
\newcommand{\YearMax}{2027}
\newcommand{\YearMedian}{2007}
\newcommand{\GenreShare}{89\%}

\newcommand{\DBPercentile}{20}
\newcommand{\DBMin}{4}
\newcommand{\CohBar}{0.8}

\newcommand{\MinSplit}{8}

\newcommand{\NegRandom}{2{,}000}
\newcommand{\NegPool}{4{,}000}
\newcommand{\NegHard}{1{,}000}

\newcommand{\NBg}{25{,}000}
\newcommand{\NBgHalf}{12{,}500}
\newcommand{\NNegWin}{40{,}000}

\newcommand{\GrowCap}{4{,}000}
\newcommand{\NWinClusters}{2{,}236}

\newcommand{\NProbes}{1{,}221}
\newcommand{\NGroupsIn}{2{,}076}

\newcommand{\TsneKeptMean}{2.5}

\newcommand{\NSubtropes}{14}
\newcommand{\NGTPosters}{1{,}444}

\newcommand{\NSubtropePosters}{269}

\newcommand{\NicheMinSize}{10}
\newcommand{\NichePlaced}{1{,}819}

\newcommand{\RhoSizeDisc}{-0.24}

\newcommand{\NDiscRanked}{31{,}834}
\newcommand{\DiscMinSize}{10}
\newcommand{\NDiscCovers}{4{,}671{,}234}
\newcommand{\DiscName}{DISCOGS-4.7M\xspace}
\newcommand{\DiscYearMin}{1860}
\newcommand{\DiscYearMax}{2026}
\newcommand{\NDiscAll}{6{,}244{,}375}

\newcommand{\DiscDesignBar}{0.94}
\newcommand{\DiscJudged}{720}

\newcommand{\NDiscWindows}{60{,}726{,}042}
\newcommand{\NDiscSeeds}{119{,}285}
\newcommand{\NDiscDet}{89{,}134}
\newcommand{\NDiscDetCovers}{4{,}139{,}848}
\newcommand{\DiscSizeMedian}{9}
\newcommand{\NDiscAnswers}{58{,}892{,}241}
\newcommand{\NDiscDetWin}{20{,}118}
\newcommand{\NDiscDetWhole}{69{,}016}

\newcommand{\DiscCap}{4{,}000}
\newcommand{\NDiscAtCap}{1{,}645}

\newcommand{\VlmSide}{640}
\newcommand{\CapTokens}{120}
\newcommand{\NameTokens}{24}
\newcommand{\CheckTokens}{2}
\newcommand{\RooyMin}{10}
\newcommand{\SinghK}{4{,}088}

\newcommand{\SinghFullPatches}{12.1 million}
\newcommand{\SinghFullK}{3.0 million}

\newcommand{\SinghS}{1{,}044{,}719}
\newcommand{\SinghIters}{136}
\newcommand{\SinghRounds}{6}
\newcommand{\SinghTop}{5}
\newcommand{\SinghHogK}{2{,}000}
\newcommand{\SinghHogDet}{3{,}472}
\newcommand{\SinghHogPosters}{1{,}417}

\newcommand{\MergeKMin}{5}

\newcommand{\GCRoundsGrow}{20}

\newcommand{\NLineWhole}{1{,}538}

\newcommand{\GCMergedGroups}{545}
\newcommand{\GCDet}{3{,}037}

\newcommand{\GCSizeMedian}{17}
\newcommand{\GCPosters}{91{,}265}

\newcommand{\NNoDet}{69{,}477}
\newcommand{\ClustersRooy}{2{,}999}
\newcommand{\FoundRooy}{8}

\newcommand{\CohRooy}{0.67}

\newcommand{\ClustersSingh}{4{,}088}
\newcommand{\FoundSingh}{8}

\newcommand{\CohSingh}{0.78}

\newcommand{\ClustersOursPlain}{3{,}037}
\newcommand{\FoundOursPlain}{10}

\newcommand{\CohOursPlain}{0.79}

\newcommand{\DaysOursPlain}{0.31}

\newcommand{\FoundWin}{5}
\newcommand{\FoundWhole}{8}
\newcommand{\FoundOnlyWinList}{face crop and overarching legs}

\newcommand{\EyeCountries}{47}
\newcommand{\EyeShareSince}{81\%}

\newcommand{\EyeOrderShort}{France (1926), Poland (1938), the US (1957) and the UK (1958)}

\newcommand{\SubtropeRows}{back-to-back & real back to back & 9 & 4 & -- & -- \\
backs & head back & 16 & -- & 5 & 11 \\
backs & two people in front of a journey & 10 & -- & -- & -- \\
eyes & center eye & 48 & 18 & 36 & 47 \\
eyes & face crop & 13 & 8 & 6 & 13 \\
eyes & half face & 14 & 6 & -- & 4 \\
eyes & portrait & 35 & 5 & 4 & 8 \\
guns & firing in the sky & 6 & -- & -- & -- \\
hands & drakula & 8 & -- & 6 & 8 \\
hands & clenched fist & 14 & -- & 11 & 12 \\
hands & hand print & 13 & 5 & 10 & 10 \\
houses & haunted house & 19 & 5 & 15 & 10 \\
legs & overarching legs & 54 & 34 & 5 & 8 \\
skulls & centered skull & 10 & -- & -- & 4 \\}
\newcommand{\PairedSingh}{8}

\newcommand{\PairedRooy}{9}
\newcommand{\PairedNotFoundRooy}{face crop}
\newcommand{\PairedOursPlain}{11}
\newcommand{\PairedNotFoundOursPlain}{half face}

\newcommand{\PairedSinghOnlyWord}{one}

\newcommand{\PairedOnlyOursList}{centered skull}

\newcommand{\RooyFaceCropHeld}{6}
\newcommand{\RooyFaceCropN}{13}
\newcommand{\PairFoundSettings}{90}
\newcommand{\PairFoundDiffer}{84}
\newcommand{\PairFoundGapMedian}{3}
\newcommand{\PairFoundGapMax}{6}
\newcommand{\RooyFaceCropLabeled}{21}

\makeatletter
\@namedef{lift@1920}{30}\@namedef{liftin@1920}{14}\@namedef{liftknown@1920}{26}\@namedef{liftshare@1920}{2\%}\@namedef{liftmax@1920}{54.4}\@namedef{liftrank@1920}{2nd}\@namedef{lifteligible@1920}{1{,}819}
\@namedef{lift@1970}{13}\@namedef{liftin@1970}{22}\@namedef{liftknown@1970}{27}\@namedef{liftshare@1970}{6\%}\@namedef{liftmax@1970}{16.4}\@namedef{liftrank@1970}{2nd}\@namedef{lifteligible@1970}{1{,}819}
\@namedef{lift@1980}{12}\@namedef{liftin@1980}{63}\@namedef{liftknown@1980}{83}\@namedef{liftshare@1980}{6\%}\@namedef{liftmax@1980}{15.6}\@namedef{liftrank@1980}{1st}\@namedef{lifteligible@1980}{1{,}819}
\@namedef{lift@2010}{2}\@namedef{liftin@2010}{21}\@namedef{liftknown@2010}{28}\@namedef{liftshare@2010}{31\%}\@namedef{liftmax@2010}{3.3}\@namedef{liftrank@2010}{21st}\@namedef{lifteligible@2010}{1{,}819}
\@namedef{lift@Western}{32}\@namedef{liftin@Western}{93}\@namedef{liftknown@Western}{96}\@namedef{liftshare@Western}{3\%}\@namedef{liftmax@Western}{33.4}\@namedef{liftrank@Western}{11th}\@namedef{lifteligible@Western}{1{,}818}
\@namedef{lift@Science Fiction}{13}\@namedef{liftin@Science Fiction}{39}\@namedef{liftknown@Science Fiction}{66}\@namedef{liftshare@Science Fiction}{5\%}\@namedef{liftmax@Science Fiction}{21.5}\@namedef{liftrank@Science Fiction}{13th}\@namedef{lifteligible@Science Fiction}{1{,}818}
\@namedef{lift@Horror}{7}\@namedef{liftin@Horror}{113}\@namedef{liftknown@Horror}{160}\@namedef{liftshare@Horror}{10\%}\@namedef{liftmax@Horror}{9.8}\@namedef{liftrank@Horror}{23rd}\@namedef{lifteligible@Horror}{1{,}818}
\@namedef{lift@Romance}{7}\@namedef{liftin@Romance}{10}\@namedef{liftknown@Romance}{10}\@namedef{liftshare@Romance}{13\%}\@namedef{liftmax@Romance}{7.5}\@namedef{liftrank@Romance}{1st}\@namedef{lifteligible@Romance}{1{,}818}
\makeatother
\newcommand{\Lift}[2][lift]{\ifcsname #1@#2\endcsname\csname #1@#2\endcsname\else\TODO{#1 of #2: not in numbers.tex}\fi}

\newcommand{\MetaYear}{160{,}359}
\newcommand{\MetaGenre}{142{,}980}
\newcommand{\MetaCountry}{146{,}313}
\newcommand{\MetaDirector}{147{,}502}
\newcommand{\DiscMetaYear}{4{,}606{,}431}
\newcommand{\DiscMetaGenre}{4{,}664{,}746}
\newcommand{\DiscMetaStyle}{4{,}136{,}425}
\newcommand{\DiscMetaCountry}{4{,}350{,}913}
\newcommand{\DiscMetaArtist}{4{,}200{,}307}
\newcommand{\DiscMetaLabel}{4{,}664{,}746}

\newcommand{\FeatSide}{512}
\newcommand{\FeatGrid}{32}
\newcommand{\WinSmall}{16}
\newcommand{\WinLarge}{24}
\newcommand{\WinStride}{8}
\newcommand{\PcaDim}{128}
\newcommand{\PoolGrid}{4}
\newcommand{\NTopLink}{3}

\newcommand{\NName}{6}

\newcommand{\NTypical}{10}

\newcommand{\NameWordsCut}{8}

\newcommand{\CtryConcEarly}{78\%}
\newcommand{\CtryConcRandEarly}{69\%}

\newcommand{\CtryConcLate}{42\%}
\newcommand{\CtryConcRandLate}{39\%}

\newcommand{\SizeMedRooy}{18}

\newcommand{\SizeMedSingh}{62}
\newcommand{\SizeMinSingh}{5}
\newcommand{\LiuDays}{22}
\newcommand{\LiuHours}{536}
\newcommand{\HoursOursPlain}{7.4}
\newcommand{\APRowMotifs}{0.39 (8) & 0.47 (9) & \textbf{0.57} (11)}
\newcommand{\APSetsMotifs}{14}
\newcommand{\APAnnotatedMotifs}{1{,}444}

\newcommand{\APSetsCourtoisChecked}{25}
\newcommand{\APAnnotatedCourtoisChecked}{366}
\newcommand{\APRowCourtoisConcept}{0.34 (14) & 0.45 (16) & \textbf{0.51} (18)}
\newcommand{\APSetsCourtoisConcept}{25}
\newcommand{\APAnnotatedCourtoisConcept}{868}
\newcommand{\APSanityMax}{0.02}

\newcommand{\NCourtoisMosaics}{34}
\newcommand{\NCourtoisTiled}{1{,}182}
\newcommand{\GTListPosters}{1{,}001}
\newcommand{\GTCategories}{12}
\newcommand{\GTAddedTool}{423}
\newcommand{\GTAddedPages}{37}
\newcommand{\GTRemoved}{17}
\newcommand{\NSubtropeFromLists}{131}
\newcommand{\ClustersSinghHogTrue}{171}
\newcommand{\ClustersRooyIncep}{2{,}856}
\newcommand{\InceptionSide}{299}
\newcommand{\SinghTrueSample}{50{,}000}
\newcommand{\SinghTrueRounds}{4}
\newcommand{\SinghTrueTop}{5}
\newcommand{\SinghTrueTopN}{40}
\newcommand{\PcaFitPatches}{120{,}000}
\newcommand{\PcaFitPosters}{2{,}000}

\newcommand{\PostersPctOurs}{56.8}
\newcommand{\PostersPctDbscan}{17.2}
\newcommand{\AblationRows}{DBSCAN & -- & 3{,}898 & 17.2 & 0.26 (4) & 0.23 (8) \\
DBSCAN & Singh & 2{,}236 & 48.4 & 0.38 (8) & 0.30 (12) \\
\cellcolor{oursbg}DBSCAN & \cellcolor{oursbg}VLM & 3{,}037 & 56.8 & 0.57 (\textbf{11}) & \textbf{0.51} (\textbf{18}) \\
$k$-means & -- & 4{,}088 & 50.0 & 0.29 (6) & 0.33 (13) \\
\cellcolor{singhbg}$k$-means & \cellcolor{singhbg}Singh & 4{,}088 & 61.1 & 0.39 (8) & 0.34 (14) \\
$k$-means & VLM & 2{,}329 & 51.9 & 0.53 (\textbf{11}) & 0.45 (15) \\
$e$-means & -- & 4{,}088 & 50.0 & 0.29 (7) & 0.32 (13) \\
$e$-means & Singh & 4{,}088 & 60.2 & 0.36 (7) & 0.33 (13) \\
$e$-means & VLM & 2{,}357 & 53.3 & \textbf{0.60} (\textbf{11}) & 0.47 (16) \\}

\newcommand{\DiscWindowsGB}{62}

\newcommand{\SinghFullDays}{18}
\newcommand{\EmbedRowsTsne}{\cellcolor{rooybg}2 & \cellcolor{rooybg}DBSCAN & \cellcolor{rooybg}-- & 2{,}999 & 69.7 & 0.47 (9) & 0.45 (16) \\
2 & DBSCAN & Singh & 2{,}999 & 52.0 & \textbf{0.54} (\textbf{12}) & 0.35 (12) \\
2 & DBSCAN & VLM & 2{,}674 & 68.6 & 0.52 (\textbf{12}) & 0.56 (\textbf{21}) \\
2 & $k$-means & -- & 1{,}603 & 100.0 & 0.53 (11) & 0.51 (17) \\
2 & $k$-means & Singh & 1{,}603 & 42.0 & 0.43 (10) & 0.40 (15) \\
2 & $k$-means & VLM & 1{,}426 & 68.0 & 0.43 (10) & 0.52 (18) \\
2 & $e$-means & -- & 1{,}604 & 100.0 & 0.47 (10) & 0.47 (16) \\
2 & $e$-means & Singh & 1{,}604 & 42.1 & 0.39 (10) & 0.37 (13) \\
2 & $e$-means & VLM & 1{,}440 & 68.0 & 0.47 (11) & 0.50 (17) \\
5 & DBSCAN & -- & 2{,}920 & 67.4 & 0.44 (8) & 0.43 (14) \\
5 & DBSCAN & Singh & 2{,}920 & 52.2 & 0.50 (11) & 0.42 (17) \\
5 & DBSCAN & VLM & 2{,}596 & 67.3 & 0.53 (\textbf{12}) & 0.56 (19) \\
5 & $k$-means & -- & 1{,}600 & 100.0 & 0.50 (11) & 0.48 (16) \\
5 & $k$-means & Singh & 1{,}600 & 42.4 & 0.44 (9) & 0.33 (13) \\
5 & $k$-means & VLM & 1{,}453 & 68.0 & 0.51 (11) & 0.50 (18) \\
5 & $e$-means & -- & 1{,}604 & 100.0 & 0.41 (10) & 0.50 (17) \\
5 & $e$-means & Singh & 1{,}604 & 43.4 & 0.40 (9) & 0.31 (12) \\
5 & $e$-means & VLM & 1{,}476 & 69.2 & 0.42 (11) & 0.54 (19) \\
15 & DBSCAN & -- & 2{,}978 & 67.2 & 0.44 (8) & 0.43 (14) \\
15 & DBSCAN & Singh & 2{,}978 & 52.0 & 0.42 (9) & 0.36 (14) \\
15 & DBSCAN & VLM & 2{,}651 & 67.0 & 0.53 (\textbf{12}) & 0.56 (19) \\
15 & $k$-means & -- & 1{,}596 & 100.0 & 0.50 (10) & 0.50 (17) \\
15 & $k$-means & Singh & 1{,}596 & 42.9 & 0.50 (11) & 0.38 (14) \\
15 & $k$-means & VLM & 1{,}442 & 68.6 & \textbf{0.54} (\textbf{12}) & 0.50 (17) \\
15 & $e$-means & -- & 1{,}602 & 100.0 & 0.42 (8) & 0.51 (17) \\
15 & $e$-means & Singh & 1{,}602 & 43.7 & 0.43 (10) & 0.34 (12) \\
15 & $e$-means & VLM & 1{,}470 & 69.0 & 0.46 (11) & 0.54 (19) \\
50 & DBSCAN & -- & 2{,}954 & 67.3 & 0.44 (8) & 0.45 (15) \\
50 & DBSCAN & Singh & 2{,}954 & 52.1 & 0.45 (10) & 0.42 (16) \\
50 & DBSCAN & VLM & 2{,}648 & 67.5 & 0.53 (\textbf{12}) & \textbf{0.57} (20) \\
50 & $k$-means & -- & 1{,}596 & 100.0 & 0.48 (9) & 0.50 (17) \\
50 & $k$-means & Singh & 1{,}596 & 42.7 & 0.43 (9) & 0.38 (14) \\
50 & $k$-means & VLM & 1{,}447 & 69.2 & 0.52 (11) & 0.49 (17) \\
50 & $e$-means & -- & 1{,}603 & 100.0 & 0.38 (9) & 0.47 (16) \\
50 & $e$-means & Singh & 1{,}603 & 42.9 & 0.38 (9) & 0.35 (13) \\
50 & $e$-means & VLM & 1{,}462 & 69.9 & 0.39 (10) & 0.51 (18) \\}
\newcommand{\EmbedRowsUmap}{2 & DBSCAN & -- & 1{,}636 & 70.7 & 0.21 (6) & 0.24 (11) \\
2 & DBSCAN & Singh & 1{,}636 & 36.2 & 0.34 (7) & 0.27 (9) \\
2 & DBSCAN & VLM & 1{,}288 & 54.8 & 0.25 (8) & 0.47 (\textbf{19}) \\
2 & $k$-means & -- & 1{,}603 & 100.0 & 0.30 (7) & 0.36 (12) \\
2 & $k$-means & Singh & 1{,}603 & 35.8 & 0.30 (6) & 0.35 (13) \\
2 & $k$-means & VLM & 1{,}181 & 61.6 & 0.46 (11) & 0.46 (17) \\
2 & $e$-means & -- & 1{,}600 & 100.0 & 0.25 (5) & 0.36 (12) \\
2 & $e$-means & Singh & 1{,}600 & 36.5 & 0.37 (8) & 0.36 (14) \\
2 & $e$-means & VLM & 1{,}179 & 62.6 & 0.34 (9) & 0.48 (16) \\
5 & DBSCAN & -- & 664 & 63.1 & 0.18 (6) & 0.26 (11) \\
5 & DBSCAN & Singh & 664 & 26.8 & 0.35 (7) & 0.22 (7) \\
5 & DBSCAN & VLM & 634 & 36.3 & 0.25 (6) & 0.29 (10) \\
5 & $k$-means & -- & 1{,}604 & 100.0 & 0.41 (9) & 0.44 (15) \\
5 & $k$-means & Singh & 1{,}604 & 41.5 & 0.38 (9) & 0.35 (13) \\
5 & $k$-means & VLM & 1{,}373 & 66.9 & \textbf{0.50} (\textbf{12}) & 0.49 (17) \\
5 & $e$-means & -- & 1{,}604 & 100.0 & 0.36 (8) & 0.43 (14) \\
5 & $e$-means & Singh & 1{,}604 & 41.0 & 0.38 (9) & 0.33 (13) \\
5 & $e$-means & VLM & 1{,}383 & 67.5 & 0.44 (11) & \textbf{0.52} (18) \\
15 & DBSCAN & -- & 607 & 62.0 & 0.19 (7) & 0.25 (11) \\
15 & DBSCAN & Singh & 607 & 26.4 & 0.37 (9) & 0.19 (7) \\
15 & DBSCAN & VLM & 582 & 33.3 & 0.35 (8) & 0.25 (10) \\
15 & $k$-means & -- & 1{,}604 & 100.0 & 0.37 (8) & 0.43 (14) \\
15 & $k$-means & Singh & 1{,}604 & 41.3 & 0.43 (9) & 0.34 (12) \\
15 & $k$-means & VLM & 1{,}390 & 68.0 & 0.44 (11) & 0.47 (17) \\
15 & $e$-means & -- & 1{,}604 & 100.0 & 0.39 (9) & 0.46 (16) \\
15 & $e$-means & Singh & 1{,}604 & 41.0 & 0.35 (8) & 0.32 (12) \\
15 & $e$-means & VLM & 1{,}410 & 68.1 & 0.42 (11) & 0.48 (17) \\
50 & DBSCAN & -- & 643 & 62.2 & 0.23 (8) & 0.25 (11) \\
50 & DBSCAN & Singh & 643 & 27.5 & 0.38 (9) & 0.18 (6) \\
50 & DBSCAN & VLM & 618 & 31.0 & 0.29 (8) & 0.28 (10) \\
50 & $k$-means & -- & 1{,}603 & 100.0 & 0.37 (8) & 0.43 (14) \\
50 & $k$-means & Singh & 1{,}603 & 41.4 & 0.38 (8) & 0.35 (13) \\
50 & $k$-means & VLM & 1{,}386 & 67.9 & 0.44 (11) & 0.51 (\textbf{19}) \\
50 & $e$-means & -- & 1{,}604 & 100.0 & 0.35 (8) & 0.44 (15) \\
50 & $e$-means & Singh & 1{,}604 & 40.1 & 0.36 (8) & 0.34 (12) \\
50 & $e$-means & VLM & 1{,}383 & 68.7 & 0.47 (\textbf{12}) & 0.48 (17) \\}
\newcommand{\GpuDaysDiscGrow}{12.4}
\newcommand{\GpuDaysDiscGrowVlm}{10.9}
\newcommand{\GpuDaysDiscGrowRest}{1.5}

\newcommand{\DiscTimeN}{182}

\newcommand{\DiscTimeYearA}{1945}

\newcommand{\DiscTimeCountries}{17}

\newcommand{\DiscTrendLateLo}{89}
\newcommand{\DiscTrendLateHi}{94}
\newcommand{\DiscTrendOutMax}{3.8}
\newcommand{\DiscTimeOrderShort}{Canada and France (1956), the US (1959), Italy (1960) and the UK (1961)}
\newcommand{\DiscTimeSince}{69\%}
\newcommand{\DiscTimeRock}{53\%}
\newcommand{\DiscTrendHipHopMin}{1991}
\newcommand{\DiscCtryEarly}{45\%}
\newcommand{\DiscCtryLate}{41\%}
\newcommand{\DiscCtryRandEarly}{47\%}
\newcommand{\DiscCtryRandLate}{39\%}
\makeatletter
\g@addto@macro\normalsize{\advance\baselineskip by 0pt plus 0.01pt\relax}
\let\cr@makecol\@makecol
\def\@makecol{\setbox\@tempboxa\vbox{\unvcopy\@cclv}\dimen@\vsize\advance\dimen@-\ht\@tempboxa
  \ifdim\dimen@>30pt \ifx\@botlist\@empty\def\@textbottom{\vskip\z@\@plus.0001fil}\fi\fi\cr@makecol}
\renewcommand{\crfill}[1]{\vspace{#1}}
\renewcommand{\crplace}[1]{{\@tempcnta\count\@currbox\divide\@tempcnta\@xxxii\multiply\@tempcnta\@xxxii\advance\@tempcnta\sixt@@n
  \@tfor\reserved@a:=#1\do{\if\reserved@a t\advance\@tempcnta\tw@\fi\if\reserved@a b\advance\@tempcnta4\fi\if\reserved@a p\advance\@tempcnta8\fi}%
  \global\count\@currbox\@tempcnta}}
\makeatother

\usepackage{colortbl}
\definecolor{singhc}{HTML}{3B8A3E}\definecolor{rooyc}{HTML}{C08A12}\definecolor{oursc}{HTML}{2F6DB5}
\colorlet{singhbg}{singhc!25}\colorlet{rooybg}{rooyc!25}\colorlet{oursbg}{oursc!25}
\DeclareRobustCommand{\shadebox}[2]{{\setlength{\fboxsep}{1pt}\colorbox{#1}{#2}}}

\definecolor{linkblue}{rgb}{0.21,0.49,0.74}
\usepackage[pagebackref,breaklinks,colorlinks,allcolors=linkblue]{hyperref}

\makeatletter
\newcommand{\corrfootnote}[1]{{\renewcommand{\thefootnote}{\fnsymbol{footnote}}\long\def\@makefntext##1{\noindent\@makefnmark\,##1}\footnotetext[2]{#1}}}
\makeatother

\def\confName{}
\def\confYear{}

\title{Niche vs. Cliché:\\ Grounded Visual Discovery of Design Motifs at a Century Scale}

\author{
Ioannis Siglidis$^{\dagger}$ \qquad Serge Belongie\\[4pt]%
Pioneer Centre for AI,\;University of Copenhagen
\vspace{0pt}%
}

\begin{document}
\twocolumn[{%
\renewcommand\twocolumn[1][]{#1}%
\maketitle
\begin{center}
\captionsetup{type=figure}
\includegraphics[width=\textwidth,height=157.7pt,keepaspectratio]{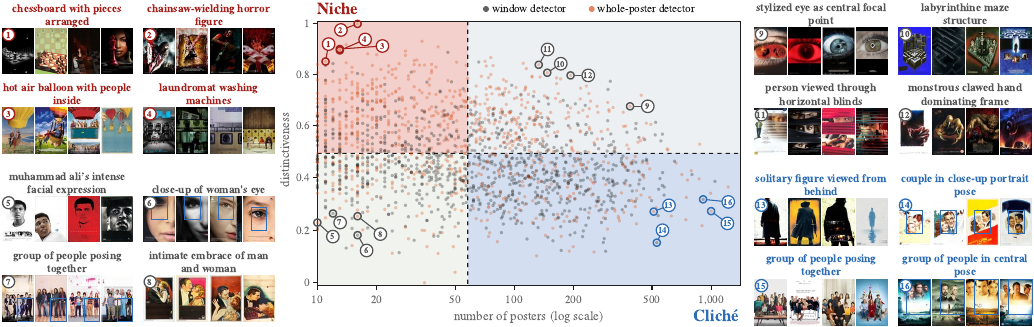}
\captionof{figure}{\textbf{Niche vs.\ cliché.} Out of a collection of $\NKept$ movie posters, each dot is a motif out of $\NichePlaced$ with at least $\NicheMinSize$ posters discovered with our approach. Each motif is arranged vertically by its relative distinctiveness (\cref{eq:distinct}) and horizontally by number of posters belonging to that motif. Gray points correspond to window detectors, while orange points correspond to whole poster detectors. We define a niche as a motif that is very distinctive but doesn't appear a lot, while cliché a motif that appears a lot but is not distinctive. See text for more details.}
\label{fig:niche}
\end{center}%
\vspace{10pt}}]
\corrfootnote{Corresponding author: {\tt ioannis.siglidis@di.ku.dk}}
\begin{abstract}
Large design collections use a repeated vocabulary of visual motifs that can appear in different contexts and styles. This paper introduces a methodology to extract grounded detectors of visual design motifs from century-long datasets of $\NKept$ posters from $\YearMin$ to $\YearMax$ and $\NDiscCovers$ album covers from $\DiscYearMin$ to $\DiscYearMax$. Our key idea is to pair a VLM-grounding strategy with standard discriminative mining~\cite{singh2012discriminative}, built on top of powerful self-supervised features. Given dense clusters provided by DBSCAN on DINOv3~\cite{simeoni2025dinov3} features, the VLM names each cluster from its most typical posters, keeps the posters that show the name, and iteratively grows a linear SVM detector, adding only posters the VLM validates. This yields $\GCDet$ named detectors holding $\GCPosters$ posters, providing better coverage and more specificity than other methods, as also evaluated on a small set of ground truth motifs. Our grounded motif detectors allow the tracking of design trends through time, genre, as well as enable the discovery of design outliers. Data, code and interactive figures are available at \url{https://niche-cliche.github.io}.
\end{abstract}
\section{Introduction}
\label{sec:intro}

\begin{figure*}[t]
\centering
\includegraphics[width=\linewidth]{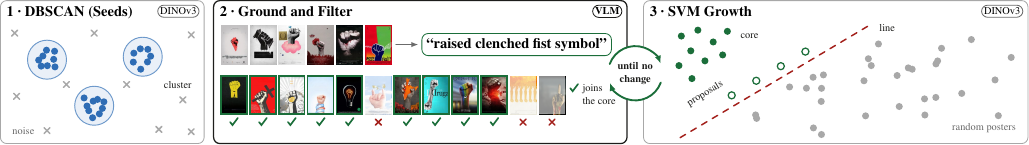}
\caption{\textbf{Approach.} \textbf{(1)} We create dense cores using DBSCAN applied on DINOv3 features. \textbf{(2)} These cores may appear diverse for example showing different types of fists, for which we ask the VLM~(Qwen3-VL-8B-Instruct~\cite{bai2025qwen3vl}) to name up to six of its most typical posters (here, ``raised clenched fist symbol''), and keep posters that belong to that in the core of the cluster. \textbf{(3)} We then use a linear SVM, trained on the core against random posters and repeat the above procedure until no new elements join the core. For more details see \cref{sec:method}.}
\label{fig:method}
\end{figure*}

Movie posters, like album covers or book jackets, or any piece of advertisement, must be both recognizable and distinctive. At a glance, one should be able to understand what the piece of media they are going to consume is about and whether they will find it interesting~\cite{haydnsmith2018selling,hipgnosis1977album,kidd2015judge}.
As a result, a compact visual vocabulary of repeated elements has been developed: an eye centered on the page, a raised fist, two characters back to back. Poster enthusiasts have documented this vocabulary by hand in blogs of clich\'es~\cite{courtois2008affiches}, wikis of tropes~\cite{tvtropes_filmposters}, letterbox lists of films~\cite{gomezrosado_lists}, or even for a specific classification in regards to a certain genre like horror~\cite{follows2017horror}. Until now, this task has been very tedious and relied on someone's personal taste and knowledge. Additionally, these classifications range in different levels of granularity, from high level cues like genre and color (e.g. ``blue documentary posters''~\cite{courtois2008affiches}), to generic devices like ``hands''~\cite{gomezrosado_lists}, or even dive deep on esoteric cultural understanding of the relationships between different devices~\cite{tvtropes_filmposters}. In this paper, our goal is to operationalize the visual discovery of a consistent visual vocabulary and to facilitate its analysis (Fig.~\ref{fig:niche}).

In 2019, a programmer and poster enthusiast called Rooy~\cite{rooy2019posters}, coded a simple pipeline that computed InceptionV3~\cite{szegedy2016rethinking} features on 169k film posters, reduced them in dimensionality using t-sne~\cite{vandermaaten2008tsne}, and then used DBSCAN~\cite{ester1996density} to cluster those posters in different motifs. While being enthusiastic about his results, Rooy notes in his blog that he spent a lot of time trying to locate interesting clusters. Drawing from Rooy's enthusiasm, our work devises a novel methodology of state of the art techniques that perform flexible discovery and grounding of visual design motifs at scale. Our idea is to start with seeds from DBSCAN on DINOV3 features, use a Vision--Language Model (VLM) to ground their common elements and then use an old method of discriminative clustering from \Singh~to grow detectors that align consistent visual appearance to semantic grounding. To our knowledge our approach is the first academic effort to computationally extract a visual vocabulary of visual motifs in either film posters or disc covers (Sec.~\ref{sec:related}). To summarize, we contribute one method and two datasets:
\begin{itemize}
    \item Grounded Visual Growth, an approach for a flexible discovery of grounded visual motifs at scale (\cref{sec:method}).
    \item \CineName: $\NKept$ film posters ($\YearMin$--$\YearMax$), with TMDB metadata (genre, country, director) and two sets of expert-annotated motifs: $\NSubtropes$ motifs among $\NGTPosters$ posters~\cite{gomezrosado_lists} and $\APSetsCourtoisChecked$ motifs over $\APAnnotatedCourtoisChecked$ posters~\cite{courtois2008affiches}.
    \item \DiscName: $\NDiscCovers$ albumDiscogs covers  ($\DiscYearMin$--$\DiscYearMax$), with metadata (genre, style, country, artist, label).
\end{itemize}

\begin{figure*}[t]
\centering
\begin{minipage}[t]{0.3165\linewidth}\centering
\includegraphics[width=\linewidth]{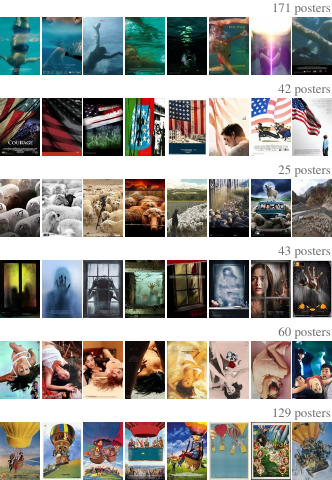}\\[2pt]
{\small(a) Singh-like~\cite{singh2012discriminative} ($\ClustersSingh$ motifs) \dotKmeans\dotSVM}
\end{minipage}\hfill
\begin{minipage}[t]{0.3165\linewidth}\centering
\includegraphics[width=\linewidth]{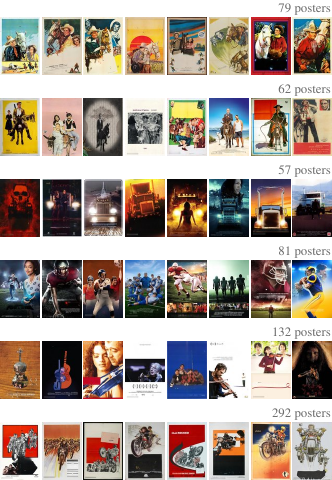}\\[2pt]
{\small(b) Rooy (2019)~\cite{rooy2019posters} ($\ClustersRooy$ motifs) \dotTSNE\dotDBSCAN }
\end{minipage}\hfill
\begin{minipage}[t]{0.3165\linewidth}\centering
\includegraphics[width=\linewidth]{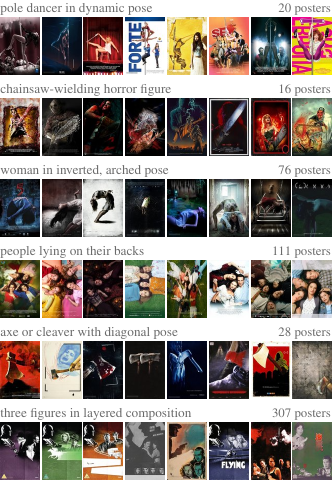}\\[2pt]
{\small(c) Ours ($\ClustersOursPlain$ motifs) \dotDBSCAN\dotSVM\dotVLM}
\end{minipage}
\caption{\textbf{Qualitative Comparison with Baselines.} For fair comparison all methods are developed using DINOv3, and are sorted by distinctiveness (\cref{eq:distinct}). Results with the original features of both approaches (InceptionV3~\cite{szegedy2016rethinking} and HOG~\cite{dalal2005histograms}), can be found in the  supplementary. Ours captures motifs all complex, grounded and visually consistent. \dotKmeans~k-means,
\dotSVM~SVM, \dotTSNE~t-SNE, \dotDBSCAN~DBSCAN,
\dotVLM~VLM.}
\label{fig:baselines}
\vspace{4pt}%
\end{figure*}

\section{Related work}
\label{sec:related}

\subsection{Image Data Mining}
\label{sec:rw_mining}
Image data mining involves a set of approaches for extracting consistent visual elements that summarize an image collection. This section discusses all the different ways previous approaches relate to our work, from supervised methods that rely on coarse labels in order to discover typical patterns, to unsupervised methods that discover elements without labels, and to VLM-grounded methods that rely on VLMs to mine important elements in the context of a dataset, without prior knowledge. We discuss prior work also in relation to film posters and disc covers, to clarify our contribution.
\myparagraph{Supervised} A strand of Image Mining approaches rely on coarse labeled sets to extract detectors from a collection of elements. Starting from images of, e.g., a city like Paris, Doersch \etal\ (2012) found the architectural details that make Paris look like Paris~\cite{doersch2012paris}, by starting with clusters of image patches and training an SVM on them while progressively selecting those that fire more in Paris and not in other cities. Later work, adapted this method using time supervision, either at certain decades~\cite{ginosar2015century}, or by finding detectors that cover a full time span~\cite{lee2013style,lee2015linking}. Supervised mining was also scaled on large and versatile image collections using diffusion models~\cite{siglidis2024diffusion}, in the cost of finetuning a generative model. Our method is both efficient and unbiased in its discovery, only relying on labels, such as genre, for analysis.

\myparagraph{Unsupervised} In fact, the approach of Doersch \etal\ (2012)~\cite{doersch2012paris} originated in \Singh which performed unsupervised discovery of mid-level patches. Starting from very dense HOG~\cite{dalal2005histograms} $k$-means clusters, it trained an SVM per cluster against a natural-world negative set, while retraining the SVM on alternating halves of the data, and keeping only the top scoring patches. Doersch \etal\ (2013)~\cite{doersch2013mid} replaced these rounds with discriminative mode seeking, that act as a form of density estimation, by adjusting the radius of each element so that a fixed mass of negative patches falls inside it. Instead, we start from dense regions of strong self-supervised features, and grow detectors grounded them on consistent semantics of the seed cluster.

\afterpage{%
\begin{figure}[t]\crplace{b}
\centering
\includegraphics[width=\linewidth,trim=0 0 0 10.0,clip]{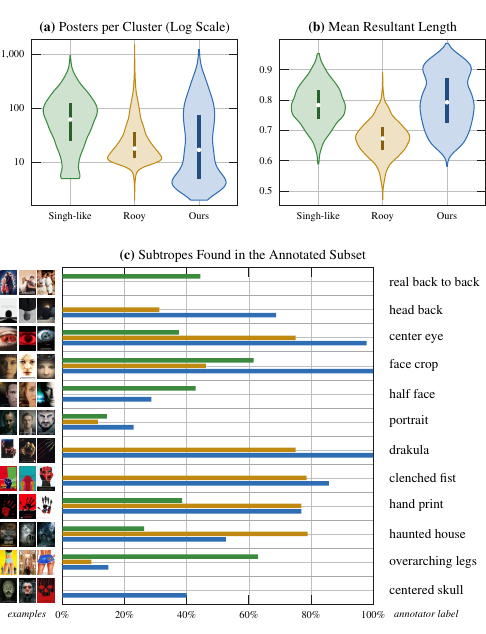}%
\caption{\textbf{Quantitative Comparison with Baselines.}  (a) Distribution of cluster size (log scale) and (b) MRL (see  \cref{sec:clusters}) of all clusters (white: median, dark: 25th to the 75th percentile). (c) Share of a motif's posters that exists in its paired cluster (\cref{tab:ap}).
}
\label{fig:baselines_stats}
\end{figure}
\begin{table}[t]
\centering
\small
\setlength{\tabcolsep}{1.6pt}
\begin{tabular}{@{}lccccc@{}}
\toprule
 Source & \#posters & \#motifs & Singh-like & Rooy & Ours \\
\midrule
Courtois~\cite{courtois2008affiches} & $\APAnnotatedCourtoisConcept$ & $\APSetsCourtoisConcept$ & \APRowCourtoisConcept \\
Letterboxd~\cite{gomezrosado_lists} & $\APAnnotatedMotifs$ & $\APSetsMotifs$ & \APRowMotifs \\

\bottomrule
\end{tabular}
\caption{\textbf{Ground Truth Motifs.} Mean average precision over the annotated motifs of each source (\cref{sec:dataset}). An annotated motif is scored by the detector that holds at least $\kmin$ of its posters and ranks them best among all annotated posters (for Rooy, we also fit a linear SVM fit on each of its clusters). In parentheses: annotated motifs held. An annotated motif that no detector holds scores 0.}
\label{tab:ap}
\end{table}
\begin{figure*}[t]
\centering
\begin{minipage}[t]{\columnwidth}
\centering
\includegraphics[width=\linewidth]{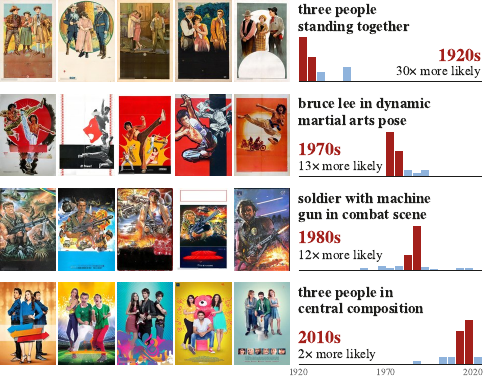}
\caption{\textbf{Typical motifs of a decade.} For four different decades, we display motif detectors of the posters that are typical of that decade. While detectors may both express similar concepts like ``three people standing together'' their visual appearance might be critical to identify them at a specific decade (top vs. bottom).}
\label{fig:period}
\end{minipage}\hfill
\begin{minipage}[t]{\columnwidth}
\centering
\includegraphics[width=\linewidth]{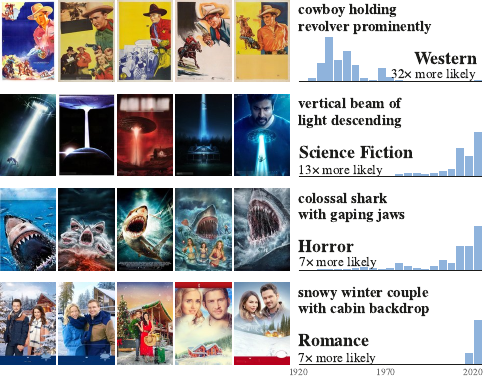}
\caption{\textbf{Typical Motifs to a genre.} We show motifs highly typical to a genre. Notice that even if labels where not used to retrieve the elements, our vlm grounding approach retrieves clusters that can be highly consistent to a specific genre.}
\label{fig:genre}
\end{minipage}
\end{figure*}
}

\myparagraph{Grounded} 
Using models that ground images into categories has been a trending recent approach for mining visual motifs in large image collection. For example, StreetStyle~\cite{matzen2017streetstyle} clusters images into $400$ styles using a CNN trained on twelve clothing attributes. More similar to us, Chiquier \etal\ (2024)~\cite{chiquier2024evolving} evolved interpretable classifiers by using an LLM to propose attributes in order organize sets of images into classes, and used CLIP~\cite{radford2021clip} to select them. Still several approaches use VLMs to perform a sequence of text aggregations: Kwon \etal\ (2024)~\cite{kwon2024ictc}, used a VLM to caption images and then asked an LLM to name $K$ clusters from the captions and assign each caption to one; Dunlap \etal\ (2024)~\cite{dunlap2024describing} compared image collections across each other by captioning images and aggregate their statistically discriminative differences; Visual Chronicles~\cite{deng2025chronicles} uses a VLM to detect what changed in each location's Street View images over time, and turned these change descriptions into trends. Most recently, to organize large image collections Liu \etal\ (2026)~\cite{liu2026organizing} uses a VLM to captions every image, then uses an LLM to propose criteria given the captions, then again a VLM to recaption the images according to the criteria, and then uses again an LLM to name the groups, assigning each image to one. Except from operating directly in the text domain~\cite{liu2026organizing} is computationally infeasible: at our scale four criteria on $5{,}000$ COCO images take $16.7$ hours on one H100, which at the same rate would take about $\LiuDays$ days ($\LiuHours$ hours) for our $\NKept$ posters, against ours $\HoursOursPlain$ hours. Unlike all previous approaches, our method is more efficient, and groups images primarily in image space, where the VLM only acts as semantic grounding, through verification.

\subsection{Visual Motifs in Posters and Album Covers}
\label{sec:rw_archives}
Unlike paintings~\cite{seguin2016link,shen2019discovering} where copying is rare and precise, or letter shapes in historical documents~\cite{siglidis2024typewriter} where variation is smooth, posters and album covers, share something akin to a visual language. This visual vocabulary is based on specific visual concepts which share elements and design styles, that designers learn to reproduce and compose, while audiences learn to recognize and interpret~\cite{rhodes2007poster,haydnsmith2018selling}.

\myparagraph{Hand Curation} Film poster enthusiasts have been collecting and archiving lists of repeating design motifs. The wiki index \url{tvtropes.com} collects esoteric film poster tropes such as the \emph{Floating Head Syndrome}~\cite{tvtropes_filmposters} and uses it to perform cultural analysis, while the film-industry data researcher Stephen Follows grouped all the layouts of $5{,}711$ horror posters into ten tropes~\cite{follows2017horror}. Closer to us, film distributor Courtois~\cite{courtois2008affiches} collected clusters arranged as grids of poster clich\'es: $26$ mosaics between 2010 and 2014 each tiling ten to about $160$ posters that share one motif clustering a total of $1{,}200$ posters. Film buf David G\'omez-Rosado on \url{letterboxd.com} curated lists of films sharing a common object like hands, legs or skulls (\eg $972$ Hand Posters)~\cite{gomezrosado_lists}. Both these catalogs demonstrate that the vocabulary we search for is real, but they rely on subjective personal preference and expertise, without describing a clear methodology of characterizing similarity. Without any explicit supervision, our approach can gather $\GCPosters$ distinct posters into $\GCDet$ motif detectors in $\HoursOursPlain$ hours. We rely on both~\cite{courtois2008affiches,gomezrosado_lists} for purposes of evaluation (\cref{sec:gteval}).  

\begin{figure*}[t]
\centering
\includegraphics[width=\linewidth]{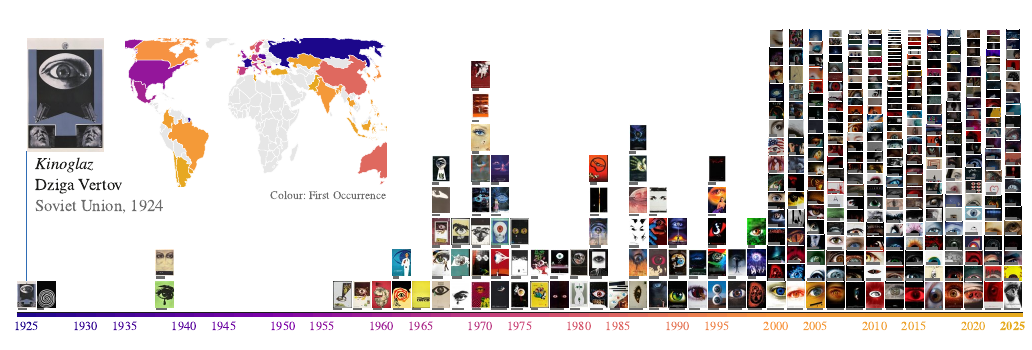}
\caption{\textbf{The centered eye through time.}  Complete posters of the ``stylized eye as central focal point'' detector arranged by release year, each column covering two years, gray bar shows confidence $\hat{s}_d(x)$ (see \cref{eq:z}) with highest predictions being at the bottom. 
}
\label{fig:time}
\end{figure*}

\noindent\textbf{In Computer Vision,} film posters have mostly been used for genre prediction task~\cite{arevalo2017gated, chu2017movie,nareti2025demystifying} or for measuring ethnic~\cite{kagan2024ethnic} and gender~\cite{song2024gender} representation with datasets ranging up to $25k$ images. MPDS~\cite{xu2026mpds}, downloaded a dataset of $373$k posters, with captions and tagged actors to train a image diffusion model, for generating film posters. Closest to us, Rooy in his 2019 blog~\cite{rooy2019posters} uses t-SNE to reduce the dimensionality of about $161$k InceptionV3 features~\cite{szegedy2016rethinking} and runs DBSCAN to aggregate results, discovering interesting results among a long set of uninteresting ones. Genre prediction has also been a task in album covers~\cite{oramas2017multilabel,dorochowicz2019relationship} with datasets of about $30$k images. Relevant to visual analysis, Fracaro \etal\ (2026)~\cite{fracaro2025disccover} computes the complexity of $46$k album covers (using different measures) showing that it rises until the mid-1990s and falls toward minimalism (after 2010). Closest to our work, Andrade L\'opez \& Flexer (2025)~\cite{andradelopez2025zeroshot} uses an image captioner to describe $3{,}130$ Billboard~200 album covers, keeps $38$ object classes from their most frequent words and counts their appearance through a segmentation network, per genre and uses metadata to show which objects (e.g. cowboy hat, skull, guitar) appear and how often on the covers of the US best-selling albums of 1945--2023. To the best of our knowledge no previous work recovers visually grounded design motifs across whole collections of film posters, or disc covers, also while using the same method.

\section{Grounded Visual Growth}
\label{sec:method}

Our approach turns an image collection into a set of named motif detectors, of a consistent visual appearance (\cref{fig:method}). First we build visually dense proposals, then ask a VLM to find elements shared across a group and then train a visual detector to be visually and semantically consistent. We first present our approach on whole posters DINOv3~\cite{simeoni2025dinov3} features but later extend it to windows, to allow a motif to be a local consistent part of a poster; all numbers are for \CineName. Implementation details and further information on all steps can be found in the supplementary.

\myparagraph{Recursive DBSCAN}\label{sec:clusters}
Motivated by Rooy (2019)~\cite{rooy2019posters} we would like to use DBSCAN to initialize our mining algorithm with a variable number of dense input proposals (\cref{fig:method}, step 1). We find that while t-SNE~\cite{vandermaaten2008tsne} produces meaningful associations (see Fig.~\ref{fig:baselines}), it doesn't preserve local neighborhoods: only $\TsneKeptMean$ of the 10 nearest of a DINOv3 ViT-L/16~\cite{simeoni2025dinov3} neighborhood of a poster are preserved when reducing dimensionality. Instead of reducing dimensionality all together what if we could train individual projections starting from well curated cores? Our approach begins by running DBSCAN~\cite{ester1996density} directly on the DINOv3~\cite{simeoni2025dinov3} features with $\kmin=\DBMin$ points per neighborhood and a radius set to the $\DBPercentile$th percentile of the $\kmin$-th neighbor distances, a value that guarantees that one poster in five has $\kmin$ neighbors.

\myparagraph{Grounding a Core}\label{sec:vlm}
Dense cores gather posters that look alike, but which not necessarily show the same element, and a linear classifier grown from it can easily drift as a look. For example, as seen in step 2 of \cref{fig:method}, a dense cluster of raised clenched fists also holds posters that share different gestures, and an SVM grown from all of them would drift towards something more generic. To this end we ground a cluster to a name, and keep only the posters that show the name. We use Qwen3-VL-8B-Instruct~\cite{bai2025qwen3vl} to first captions each of the cluster's of up to six main images separately and then names, from these six captions, what all of them share, in two to six words: an object, a figure, a pose, a scene or a graphic design, and how it appears. Intentionally, we excluded styles and genres, which are properties of a whole poster rather than elements it shows. The name is verified on all six, answering yes or no, and the model iterates this process until a name is valid for all, otherwise keeping the name that holds for the most. 

\myparagraph{Visual Growth}\label{sec:growth}
From each core we build a motif detector. Similar to \Singh, each detector $d$ is a linear SVM~\cite{cortes1995support}, defined as a function:
\begin{equation}
s_d(x) = w_d^\top x + b_d\label{eq:score}
\end{equation}
where $x\in \mathbb{R}^{D}$ is the unit-normalized features of a poster, $w_d \in \mathbb{R}^{D}$ is a linear projection vector, and $b_d$ is a bias a scalar ($D=1024$). Given a random held out set of the corpus of $\NBg$ posters, cut in two halves, a detector is trained on its core (\cref{fig:method}, step 3) against $\NegRandom$ random negatives from one half, with one round of hard-negative mining of $\NegHard$ among $\NegPool$ as in Singh \etal\ (2012)~\cite{singh2012discriminative}.
We use the other half of the held out set to define a decision boundary being the highest score it gives to the first $\NLineWhole$ posters of that half. Posters above the line are validated by the VLM until the negatives outnumber the positives. Confirmed posters join the core, and the whole process is repeated, where an SVM is fit again until no new poster is confirmed ($\GCRoundsGrow$ rounds). To avoid detectors converging to the same element we merge and refit the SVM when at least half of their first $\NTypical$ elements, closest to their centroid or central poster, are the same ($\GCMergedGroups$ merges). This results in a total of $\GCDet$ detectors of $\GCPosters$ posters.
\begin{figure*}[t]
\centering
\includegraphics[width=\linewidth]{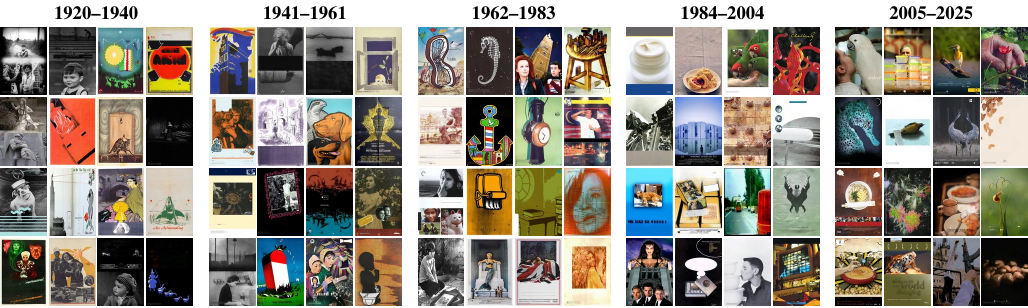}
\caption{\textbf{Outliers through time.} We can identify unique visual motifs by selecting the posters farthest from all motif detectors (\cref{eq:outlier}).}
\label{fig:outliers}
\end{figure*}
\begin{figure}[t]
\centering
\includegraphics[width=\linewidth]{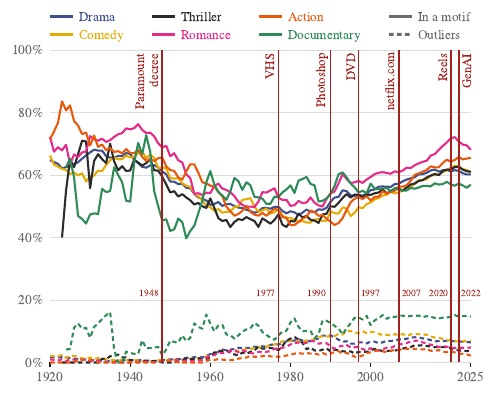}
\caption{\textbf{Motifs and outliers by genre through time.} For the six most common TMDB genres (\cref{sec:dataset}), we annotate the yearly share of posters held by a visual motif detector
(solid) and of outliers (dashed). Vertical lines signify important historical events for the movie industry, related with poster production, as well as movie production and distribution. See text for details.}
\label{fig:outliers_curve}
\end{figure}

\myparagraph{Window Proposals}\label{sec:windows}\label{sec:scales}
So far a poster is described by one vector, the CLS token of DINOv3~\cite{simeoni2025dinov3} which corresponds to the whole image. Similar to Singh \etal (2012)~\cite{singh2012discriminative} our method also incorporates window proposals, by using local DINOv3 tokens, that also provide contextual information to the scene. This requires a series of adaptations. Each poster is cut into $\WinPerPoster$ overlapping square windows at two resolutions, and two locations where its DINOv3 patch tokens are pooled and compressed with PCA to a $2{,}048$-d vector. Recursive DBSCAN then runs on the windows exactly as on the whole posters. To make a window core each poster of a cluster keeps its window closest to the cluster's mean, as otherwise nearby windows can inflate the density. To avoid duplicates, we first merge clusters that hold each other's $\NTopLink$ most typical posters, and then each core claims a poster if it has maximal similarity to its mean, resulting in $\NWinClusters$ window cores. To see if a window core add something more than a full image core we train a linear SVM on both. The window detector negatives are $\NNegWin$ windows, randomly drawn to be the same for every detector and what changes from round to round is the core and, with it, the $\NegHard$ hard negatives that the detector picks among the $\NegPool$. When the whole poster cluster has at least $\kmin$ posters and fewer than half of them are in the window cluster, these posters form an additional whole-poster proposal. $\NProbes$ of the $\NGroupsIn$ whole-poster proposals come from windows. As discussed in ~\cref{sec:analysis} having both detectors improves results.

\section{Mining Design Motifs at a Century Scale}
\label{sec:results}

\begin{figure*}[t]
\centering
\begin{minipage}[t]{0.49\linewidth}\centering
\includegraphics[width=\linewidth]{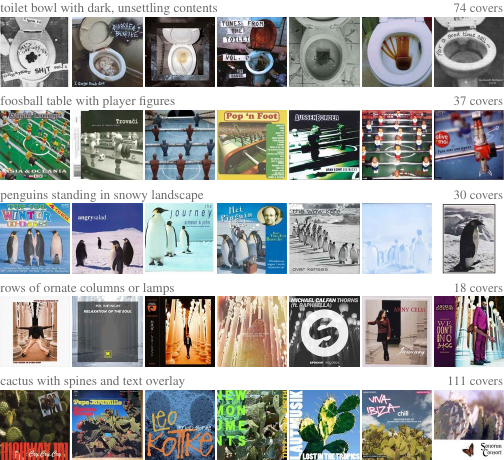}\\[2pt]
{\small(a) Most distinctive}
\end{minipage}\hfill
\begin{minipage}[t]{0.49\linewidth}\centering
\includegraphics[width=\linewidth]{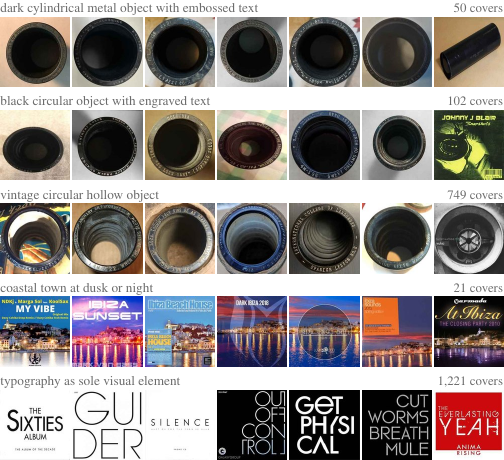}\\[2pt]
{\small(b) Least distinctive}
\end{minipage}
\caption{\textbf{Album covers.} The most \textbf{(a)} and  the least \textbf{(b)} distinctive of the $\NDiscRanked$ detectors with at least
$\DiscMinSize$ covers. Our method can scale reliably to find motifs in a much larger and more diverse dataset of disc covers. Extensive results can be found in the supplementary.}
\label{fig:discs_extremes}
\end{figure*}

\subsection{Datasets}
\label{sec:dataset}
Our work introduces two datasets, each a snapshot of an online archive at the time it was collected: \CineName, with film posters from \url{cinematerial.com} as of February 2026, and \DiscName, of album covers from \url{discogs.com} as of May 2026. Both datasets will be shared as url links and image hashes (as in LAION~\cite{schuhmann2021laion}).
\myparagraph[]{\CineName}
We scrape all primary posters of the film poster archive \url{cinematerial.com} along side their film title and release year. Since lettering can give up part of the visual grouping  or create false associations, we detect text with HunyuanOCR~\cite{hunyuanocr2025} and inpaint it with LaMa~\cite{suvorov2022lama}. Removing blank posters,
title cards and near-copies leaves $\NKept$ posters from $\YearMin$ to $\YearMax$ (median $\YearMedian$). Using a film's title and release year we identify the genre and country of origin for $\GenreShare$ of the posters. To provide an annotated subset we start from Letterboxd~\cite{gomezrosado_lists}, that each gather films whose posters share one device (e.g. hands, eyes, etc). Matching their films to our posters by title and release year gives $\NGTPosters$ annotated posters, for which we then annotate a small random holdout subset of $\NSubtropes$ visual motifs using neighboring proposals. To provide alternative groupings, we use $\NCourtoisMosaics$ Courtois mosaics~\cite{courtois2008affiches}, which we cut into their $\NCourtoisTiled$ posters, which after matching to ours and adding nearby posters that match the same motif result in $\APAnnotatedCourtoisConcept$ posters. Such catalogs are particular in what they count as motifs: Courtois's sets are ``blue posters for animal documentaries'', or ``black-and-white posters with a red title''. Our method uses neither genre nor a poster's overall color, so while we keep only those that involve visual information, we don't expect our method to find all of them.

\myparagraph[]{\DiscName}
Using \url{discogs.com} we also collected $\NDiscAll$ album covers, alongside their release year, genres, artists and labels. Since disc covers contain semantically more duplicates, as a label or an artist reissues one design, we keep only one cover of each design for each artist and each label. We cover pairs with cosine similarity DINOv3~\cite{simeoni2025dinov3} above $\DiscDesignBar$, to be duplicates, resulting is $\NDiscCovers$ covers from $\DiscYearMin$ to $\DiscYearMax$. For simplicity we don't inpaint their text, as after excluding neighbors of the same artist and the label, we don't expect text to provide grouping information.

\subsection{Analysis}
\label{sec:gteval}\label{sec:analysis}

\myparagraph{Comparison to Prior Work}\label{sec:measures}
To ablate our approach, in \cref{fig:baselines_stats} we compare it with the two methods on which it is based, \Rooy, which uses DBSCAN on a t-SNE map of whole posters ($\ClustersRooy$ clusters) and an approach like \Singh (as the original one is infeasible on our data), which applies k-means and then discriminative mining ($\ClustersSingh$ detectors). For Rooy's we only use whole posters and for Singh-like only windows. Our distribution of detectors range from a few posters to hundreds (median $\GCSizeMedian$), like Rooy's (median $\SizeMedRooy$), while Singh-like ones start at $\SizeMinSingh$ (median $\SizeMedSingh$); their MRL is as high as Singh-like (median $\CohOursPlain$ and $\CohSingh$) and higher than Rooy's ($\CohRooy$). Of the $\NSubtropes$ annotated motifs, counting a motif as found by the cluster paired with it in \cref{tab:ap}, %
the clusters of \Rooy\ find $\PairedRooy$ and the Singh-like detectors find $\PairedSingh$, while ours find $\PairedOursPlain$. We find every motif Rooy's find, all but \PairedSinghOnlyWord{} of Singh-like, and \PairedOnlyOursList{}, which neither finds. A stricter rule, also asking the motif's posters in that cluster to be half or more of its labeled ones, gives $\FoundRooy$, $\FoundSingh$ and $\FoundOursPlain$ (supplementary). Our motifs are superior in terms of Mean Average Precision in both datasets (\cref{tab:ap}).

\myparagraph{Niche vs. cliché} To create a comparable standardized detector, we use the score $c_d$ at which the growth of $d$ stopped (the last poster the VLM approved) as a \emph{cut-off} and standardize by their mean $\mu_d$ and standard deviation $\sigma_d$ over the $\NBgHalf$ held-out posters,
\begin{equation}
\hat{s}_d(x) = \frac{s_d(x) - \mu_d}{\sigma_d}.\label{eq:z}
\end{equation}
Our teaser~\cref{fig:niche} arranges $\NichePlaced$ detectors $d$ which have at least $\NicheMinSize$ elements by log size and distinctiveness,
\begin{equation}
\Delta_d = -\frac{1}{100} \sum_{x \in d_{1:10}} \sum_{d' \in \operatorname{top}_{10}(x)} \bigl(\hat{s}_{d'}(x) - c_{d'}\bigr),\label{eq:distinct}
\end{equation}
where $\mathcal{D}$ is the set of all our detectors, $d_{1:10}$ the ten (to safely avoid outliers) most typical posters of $d$, $\hat{s}_{d'}(x)$ the standardized score of \cref{eq:z} and $c_{d'}$ the cut-off where the growth of $d'$ stopped, in the same units. $\hat{s}_{d'}(x) - c_{d'} > 0$ means that $d'$ ranks $x$ above where its own growth stopped, so a detector is distinctive ($\Delta_d > 0$) when the ten other detectors that rank each of its typical posters $x$ highest, denoted as $\operatorname{top}_{10}(x)$, still rank them, on average, below their cut-offs. As we only use 10 posters, $\Delta_d$ does not grow with the detector's size. Naturally size and distinctiveness are weakly correlated (Spearman $\rho = \RhoSizeDisc$), which we define as the niche cliche axis. We define motifs to be niche when they are highly distinctive yet carry a small number of elements. On the other hand motifs are cliché when they are not distinctive but carry a high number of elements. Niche motifs include: chessboards, hot air balloons, and laundromats. Clichés motifs include: a solitary figure seen from behind, a couple in a close-up portrait pose, or a group of people posing together. We also locate motifs highly distinctive but with lots of posters such as the centered, eye or the vampire hand. Other motifs are not distinctive but have a low count, like Muhammad Ali's intense expression, or the ``intimate embrace of a man and a woman''. Here visual appearance is critical to define a motif, as the intimate embrace is part of an old cinema poster.

\begin{figure*}[t]
\centering
\begin{minipage}[t]{0.3175\linewidth}\centering
\includegraphics[width=\linewidth]{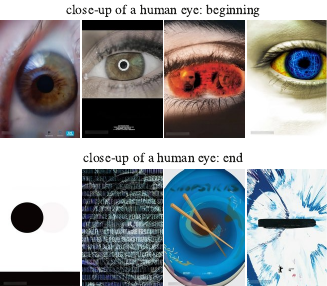}\\[2pt]
{\small(a) Precision}
\end{minipage}\hfill
\begin{minipage}[t]{0.3175\linewidth}\centering
\includegraphics[width=\linewidth]{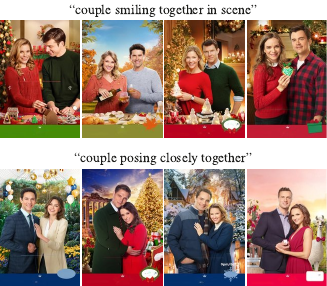}\\[2pt]
{\small(b) Fragmentation}
\end{minipage}\hfill
\begin{minipage}[t]{0.3175\linewidth}\centering
\includegraphics[width=\linewidth]{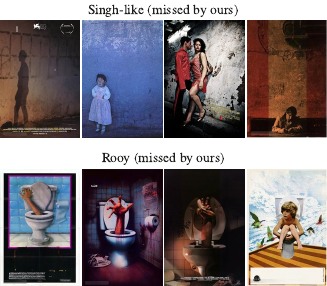}\\[2pt]
{\small(c) Completeness}
\end{minipage}
\caption{\textbf{Limitations.} (a) Precision: the tail of the detector drifts in concept cohesion. (b) Fragmentation: different detectors for a similar concept. (c) Completeness: baselines detect motifs with high MRL that our approach doesn't detect.}
\label{fig:limits}
\end{figure*}

\myparagraph{Typical Motifs across Time and Genre}\label{sec:time}
Given that most of our posters are both dated and identified in regards to genre (\cref{sec:dataset}) we can ask how typical is a motif is tied to a decade or a genre $y$. If $P(y)$ is the share of the posters of $y$ in the corpus and $P(y \mid d)$ their share in $d$, we define typicality as the ratio,
\begin{equation}
\begin{aligned}
T(y \mid d) &= \exp \mathrm{PMI}(d, y) = \frac{P(y \mid d)}{P(y)},
\end{aligned}
\end{equation}
where PMI is the pointwise mutual information. $T(y \mid d)=3$ means that the posters of $d$ belong to $y$ three times as often as posters in general. $P(y \mid d)$ sums the rectified $\hat{s}_d(x)$ (\cref{eq:z}) of the posters of $d$ that belong to $y$ and divides by the same sum over all the posters of $d$, so that the most typical posters of $d$ count the most.
In \cref{fig:period} we select 4 detectors that are associated to their decade with high probability: three people standing together during the 1920s ($\Lift{1920}\times$) and the 2010s ($\Lift{2010}\times$) with different appearance, as well as action posters of Bruce Lee in the 1970s ($\Lift{1970}\times$), and Commando movie posters in the 1980s ($\Lift{1980}\times$). We also associate detectors that are with high probability associated with their genre (\cref{fig:genre}) like a cowboy holding a revolver for Western ($\Lift{Western}\times$), a shark's open jaws  $\Lift{Horror}\times$ for horror, a beam of light descending from a UFO for sci-fi ($\Lift{Science Fiction}\times$), and a smiling couple posing together for romance ($\Lift{Romance}\times$). 
\myparagraph{Tracing a Motif's History} Using our motifs we can trace their evolution through time, for example the centered eye (\cref{fig:time}). We first detect it in Dziga Vertov's \emph{Kinoglaz} (Soviet Union, 1924), followed by \EyeOrderShort, and it becomes a very common motif during the 2000s ($\EyeShareSince$ of posters), spanning $\EyeCountries$ countries, half of which come from the US.

\myparagraph{Outliers} Given that our motif detectors can find repeating visual patterns they can also help us identify designs through time. For $\NNoDet$ posters that don't belong to any detector,
we compute an outlier score:
\begin{equation}
o(x) = 1 - \max_{d \in \mathcal{D}} \max_{q \in d}\, x^\top q,\label{eq:outlier},
\end{equation}
which measures that maximal distance to any element that belongs to a detector.
In \cref{fig:outliers} we collect outliers with the lowest similarity for each decade, like a magenta poster with a mage and ducks between the 1920--1940s, a stool made of wooden buildings between the 1962--1983, and a person occluded by tapperware between the 1984--2004.

\myparagraph{Global Trends} Alongside tracking outliers we can also track global trends \cref{fig:outliers_curve}. For each genre, each curve carries the percentage of posters carried by a detector per year averaged in a 5 year window. Before the 1950s we find the most motif dense region, which after the end of the studio system with the Paramount decree of 1948 reaches its lowest during 1970s and 1980s. Alongside the invention of VHS (1980s), Photoshop and DVDs (1990s), and the rise of netflix.com we see a general tendency of film posters to become more and more motif like, and this time at a global scale: a motif's top country holds a median $\CtryConcEarly$ of ten of its posters before 1940 vs $\CtryConcLate$ after 2005 ($\CtryConcRandEarly$ vs.\ $\CtryConcRandLate$ for random ones).

\myparagraph{Album Covers}
Our method is not specific to film posters, but works equally well to a dataset of $\NDiscCovers$ album covers which we introduced in Sec.~\ref{sec:dataset}. Running our method on album covers with a coarser window descriptor ($512$-d instead of $2{,}048$-d) yields $\NDiscDet$ detectors, with the most distinctive showing a toilet bowl, a foosball table or penguins, and the least distinctive showing old metal cylinders records or bare typography covers (\cref{fig:discs_extremes}). Extensive results on album covers can be found in the supplementary.

\myparagraph{Limitations}\label{sec:limits} While our approach does a significant step towards capturing visual grounded motifs in design collections such as film posters or disc covers it is far from perfect. The key limitations (\cref{fig:limits}) of our method include: a) that elements far from the detector can diverge to include various senses of visual appearance; b) different detectors may describe the same concept hold similar names and diverge slightly; c) baseline methods may discover salient motifs that our approach might fail to discover. \cref{fig:limits}c shows examples of clusters detected by Rooy and Singh-like whose top-10 elements don't belong to any of our detectors.

\vspace{0.3em}
\section{Conclusion}
\label{sec:conclusion}
We presented a method that discovers visually and semantically grounded design motifs on a collection of film poster and album covers. Starting from clusters of DBSCAN, we refine proposals through a VLM that names their common element, growing a linear SVM detector from those that validate it. Our approach recovers more annotated motifs than previous approaches, while grounding them to semantically consistent clusters, which aid informative investigation through time, genre, and outliers. We hope our work inspires more research automating the organization of visual culture.

\pagebreak   %
\paragraph{LLM Usage Statement}
Claude Opus 5.5 (Max) was used throughout this work as a research assistant for collecting data, related work, planning and running experiments, making figures, and writing the text. All content in this text, has been read, edited, and approved by its authors. Having a coding agent, allowed us to perform better research by helping us create multiple bespoke websites to understand deeper, navigate, and improve our results throughout all this process, as well as helped us fetch all the necessary quantitative information related to different outputs of our method, which would be infeasible to organize and collect otherwise.

\section*{Acknowledgments}
We thank Stella Frank, Noah Zegna Rothenberger and Vésteinn Snæbjarnarson for useful discussions. This project was enabled by P1-DTU-HPC Compute~\cite{DTU_DCC_resource}.

\FloatBarrier   %
{
    \small
    \bibliographystyle{ieeenat_fullname}
    \bibliography{main}
}

\newif\ifsupmat
\supmattrue
\ifsupmat
\clearpage
\setcounter{page}{1}
\maketitlesupplementary
\renewcommand{\thefigure}{S\arabic{figure}}
\setcounter{figure}{0}
\renewcommand{\thetable}{S\arabic{table}}
\setcounter{table}{0}
\renewcommand{\thesection}{\Alph{section}}
\setcounter{section}{0}
\makeatletter   %
\setlength{\@dblfptop}{0pt}\setlength{\@dblfpsep}{12pt plus 2pt}\setlength{\@dblfpbot}{0pt plus 1fil}
\setlength{\@fptop}{0pt}\setlength{\@fpsep}{12pt plus 2pt}\setlength{\@fpbot}{0pt plus 1fil}
\makeatother
\definecolor{figblue}{HTML}{2F6DB5}\definecolor{figred}{HTML}{A3201B}   %
This supplementary material provides additional details to the main part of our paper.  First, we summarize our approach formally as an algorithm, detail the construction of features, the prompt we provide to the vision--language model (\cref{sec:supp_method}). Second, we describe \CineName: its metadata, how its motifs were annotated, and how we evaluate that a method locates them (\cref{sec:supp_data}). Third, we run the two baselines as they were published, on their own features, and cross their components with ours (\cref{sec:supp_baselines}). Finally, we detail how we built \DiscName and show further results on its album covers (\cref{sec:supp_discs}).

\begin{algorithm}[t]
\caption{\textbf{Grounded Visual Growth}}
\label{alg:method}
\algrenewcommand\algorithmicrequire{\textbf{Input:}}\algrenewcommand\algorithmicensure{\textbf{Output:}}
\begin{algorithmic}[1]
\Require DINOv3 features of posters $X_p$ and windows $X_w$.
\Ensure detectors $\mathcal{D}$: SVM $d_c$, name $t_c$, posters $c$.
\Function{RecursiveDBSCAN}{$X$}
    \State $\mathcal{C} \gets \textsc{Cores}(\textsc{DBSCAN}(X))$
    \State $\mathcal{R} \gets \emptyset$
    \For{$c \in \mathcal{C}$}
        \If{$\mathrm{MRL}(c) < \tsplit \land |c| \geq \MinSplit$}
            \State $\mathcal{R} \gets \mathcal{R} \cup \Call{RecursiveDBSCAN}{c}$
        \Else
            \State $\mathcal{R} \gets \mathcal{R} \cup \{c\}$
        \EndIf
    \EndFor
    \State \Return $\mathcal{R}$
\EndFunction
\State $\mathcal{C}_w \gets \textsc{Claim}(\textsc{Merge}(\textsc{RecursiveDBSCAN}(X_w)))$
\State $\mathcal{C}_p \gets \textsc{RecursiveDBSCAN}(X_p)$
\For{$c \in \mathcal{C}_w$}
\State $d^w_c \gets \textsc{SVM}(c)$
\State $d^p_c \gets \textsc{SVM}(\textsc{Posters}(c))$
\State $g_c \gets d^p_c \setminus \textsc{Posters}(c)$
\EndFor
\State $\mathcal{C}_p \gets \mathcal{C}_p \cup \{g_c : |g_c| \geq \kmin,\ |g_c \cap d^w_c| < |g_c|/2\}$
\State $\mathcal{C}_p \gets \textsc{Claim}(\textsc{Merge}(\mathcal{C}_p))$
\For{each core $c \in \mathcal{C}_w \cup \mathcal{C}_p$}\Comment{Grounding a Core}
\State $t_c \gets \textsc{Name}(c)$
\State $c \gets \{x \in c : \textsc{Verify}(x, t_c)\}$
\EndFor
\State $\mathcal{C} \gets \{c \in \mathcal{C}_w \cup \mathcal{C}_p : c \neq \emptyset\}$
\Repeat\Comment{Visual Growth}
\For{each core $c \in \mathcal{C}$}
\Repeat
\State $d_c \gets \textsc{SVM}(c)$
\State $N \gets \textsc{Validate}(d_c, t_c) \setminus c$
\State $c \gets c \cup N$
\Until{$N = \emptyset$}
\EndFor
\State $\mathcal{C} \gets \textsc{Merge}(\mathcal{C})$
\Until{nothing merges}
\State \Return $\mathcal{D} = \{(d_c, t_c, c) : c \in \mathcal{C}\}$
\end{algorithmic}
\end{algorithm}

\section{Method details}
\label{sec:supp_method}

\begin{figure}[t]
\centering
\includegraphics[width=0.9\linewidth]{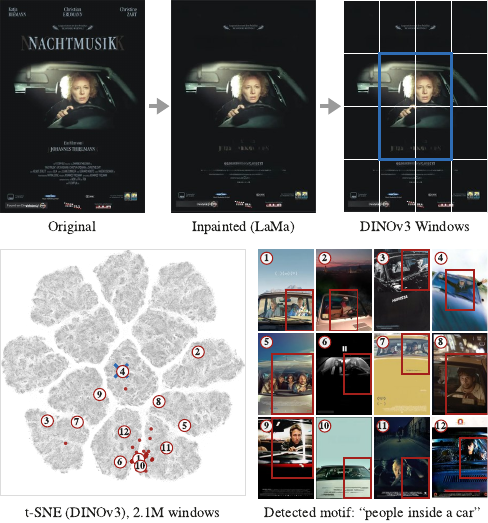}
\caption{\textbf{Data, and why no t-SNE.} Top: a poster, its inpainted version and its $\WinPerPoster$ DINOv3 windows; we
follow the \textcolor{figblue}{query window}.  Bottom left: the t-SNE map of all windows, with the query window as a
\textcolor{figblue}{cross} and the \textcolor{figred}{window dots} of a motif our method detects, ``people inside a car'', shown on the bottom right.}
\label{fig:data}
\end{figure}

\begin{figure*}[t]\crplace{p}
\centering
\includegraphics[width=1.0\linewidth]{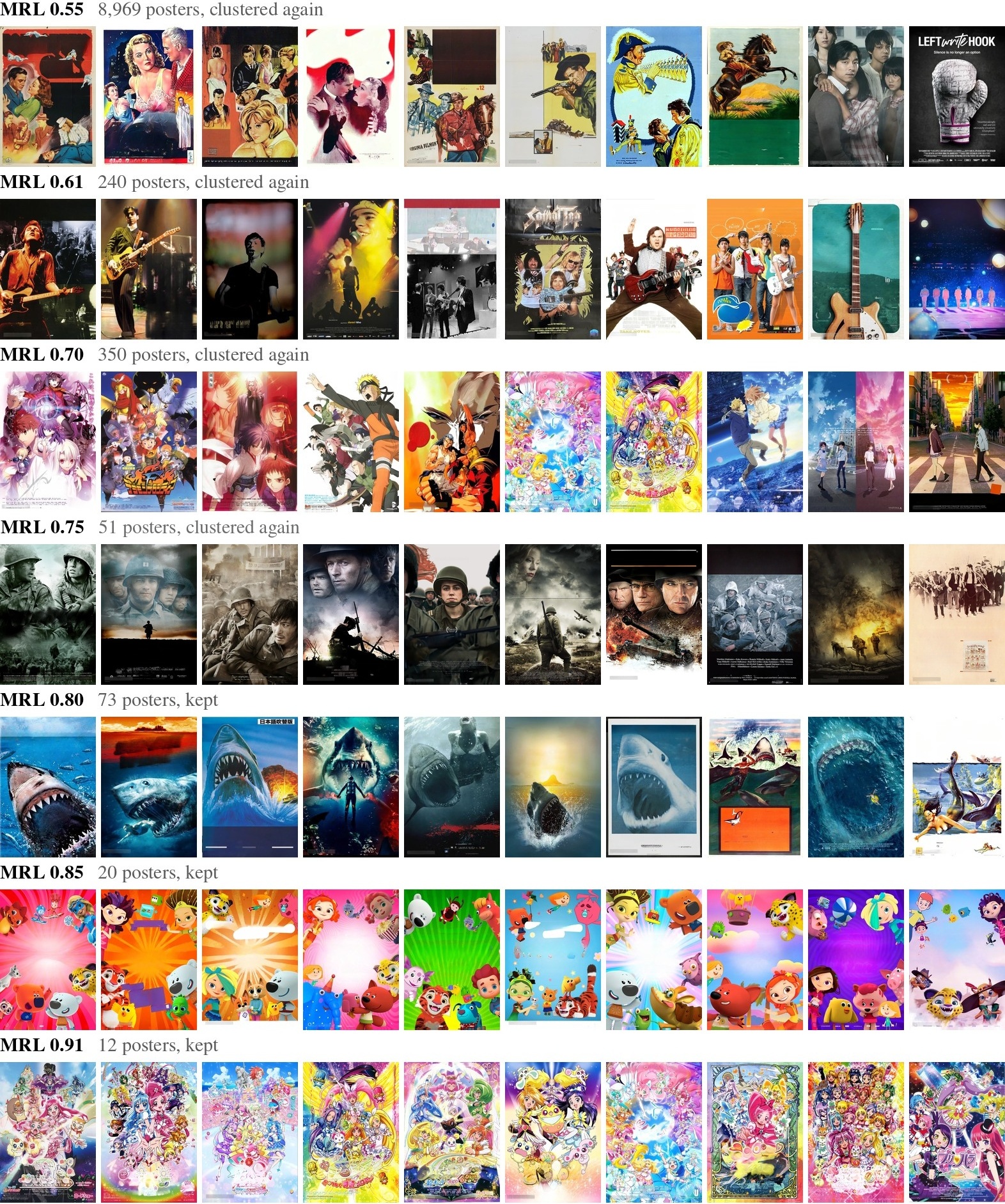}
\crfill{2.84pt}\caption{\textbf{Clusters of increasing MRL.}  Whole-poster DBSCAN clusters (\cref{sec:clusters}), each shown with ten of its posters evenly spaced from the most to the least central, \ie, from the closest to the farthest from the cluster's mean (\cref{sec:supp_method}).
The clusters under the MRL threshold $\tsplit = \CohBar$ (top four rows) are clustered again, while the others are kept.}
\label{fig:supp_cohesion}
\end{figure*}

\begin{table*}[t]
\centering
\small
\caption{\textbf{VLM Prompts.} Four exact prompts for all the VLM stages, described in \cref{sec:method} and in \cref{sec:supp_vlm}.}
\begin{tabular}{@{}lp{0.86\linewidth}@{}}
\toprule
Caption & \leavevmode Describe this image in detail: the people, animals, objects, places and actions in it, and how they appear (their
pose, size and place in the picture). Two or three sentences. \\
Name & \leavevmode Below are descriptions of $\langle n\rangle$ images that an algorithm grouped together because they share one
recurring thing. $\langle$the $n$ captions, numbered, one per line$\rangle$ What do all $\langle n\rangle$ images share? Name it in 2 to 6
words, precisely: the object, figure, pose, scene or graphic design, and how it appears (not the style and not the genre). Reply with
ONLY the name. \\
Name again & \leavevmode Below are descriptions of $\langle n\rangle$ images that an algorithm grouped together because they share one
recurring thing. $\langle$the $n$ captions$\rangle$ The name ``$\langle$name$\rangle$'' does not fit image(s) $\langle$their
numbers$\rangle$. What do all $\langle n\rangle$ images share? Name it in 2 to 6 words, precisely: the object, figure, pose, scene or
graphic design, and how it appears (not the style and not the genre). Reply with ONLY the name. \\
Verify & \leavevmode Does this image show $\langle$name$\rangle$? Answer with exactly one word: yes or no. \\
\bottomrule
\end{tabular}

\label{tab:prompts}
\end{table*}

\begin{figure*}[t]
\centering
\includegraphics[width=\linewidth]{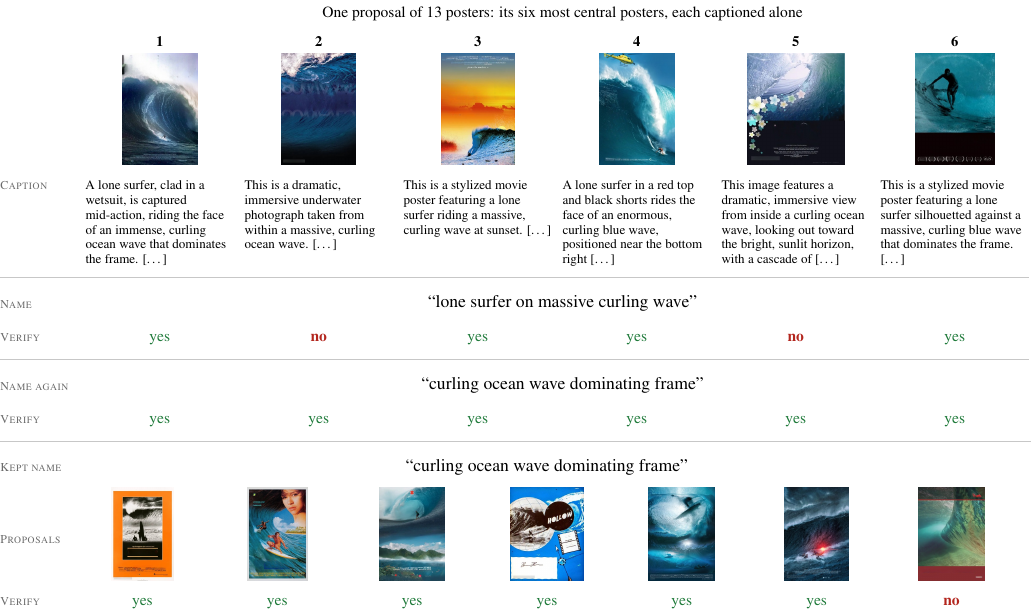}
\caption{\textbf{Example VLM outputs.} A single proposal of our run ($13$ posters) followed through the prompts of \cref{tab:prompts}. Each poster is shown as the model sees it (text inpainted), and every reply is given word for word. Its six most
central posters are captioned one by one ([\dots] marks more than one sentences). The name given from the six captions, ``lone surfer on massive curling wave'', fails on posters 2 and 5; told so, the model gives another name, ``curling ocean wave dominating frame'', which holds on all six. Asked on the proposal's other $7$ posters, the kept name holds on $6$, which join the six in the core.}
\label{fig:vlm_examples}
\end{figure*}
\cref{alg:method} summarizes our method with the steps of \cref{sec:method}. Here we give only what the main paper leaves out. Throughout, the \emph{most central} posters of a cluster or a detector (its most typical ones in the main paper) are those whose features are closest, in cosine similarity, to the mean of its members' features (their windows', for a window detector). In all our figures we show the most central
poster first, then the others closest by cosine similarity to it. In \textsc{RecursiveDBSCAN}, the mean resultant length of a cluster $c$ of unit
features $\hat{x}$ is
\begin{equation}
    \mathrm{MRL}(c) = \Big\| \frac{1}{|c|} \sum_{x \in c} \hat{x} \Big\|_2 \in [0, 1],
\end{equation}
where \cref{fig:supp_cohesion} shows the effect of its threshold on whole-poster DBSCAN clusters. The four under $\tsplit =
\CohBar$ are the ones that are clustered again. Besides dropping the noise and the small cores, Cores removes near-copies, posters whose cosine similarity to a
poster already kept is at least $\tdup$, and on windows each poster of a cluster keeps only its window closest to the cluster's mean.%
\begin{figure*}[t]
\centering
\includegraphics[width=\linewidth,height=157.7pt,keepaspectratio]{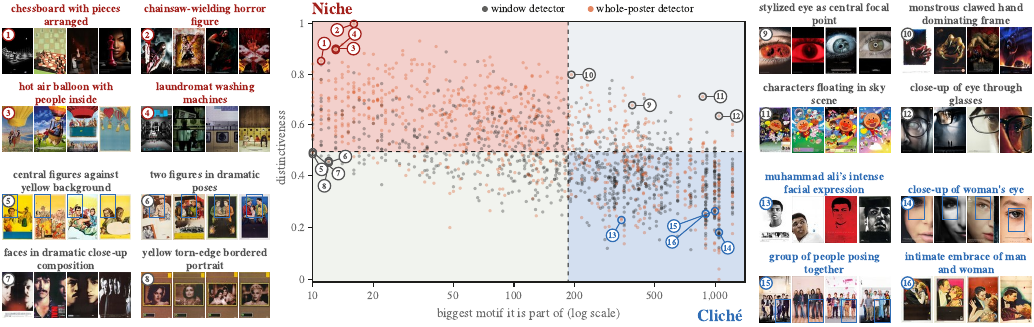}
\caption{\textbf{Niche vs.\ cliché, by the biggest motif.}  As \cref{fig:niche}, where the x-axis is the biggest motif that the elements of that detector belong to. A motif that lies inside a bigger one is thus placed at the bigger one's size.}
\label{fig:niche_vote}
\end{figure*}
Before grounding, \textsc{Merge} removes duplicate proposals, linking a proposal to another when the other contains its $\NTopLink$ most central posters, and merging the proposals of each strongly connected component of these links, in one pass. \textsc{Claim} then keeps each poster in the proposal whose mean it is closest to. We call \emph{detector} any linear SVM we train, and write $x \in d$ when $d$ scores a poster $x$ (\cref{eq:score}) at least at its decision boundary; a window detector scores $x$ by the best of its windows. Additional whole-poster proposals come from window detectors, where for each window core $c$, we train one detector on its windows, $d^w_c$, and one on its posters, $d^p_c$. The posters of $d^p_c$ outside $c$ (at most the $\GrowCap$ best) form a new whole-poster proposal $g_c$ when they are at least $\kmin$ and fewer than half of them are in $d^w_c$. Later, the detectors $d_c$ of the growth are grounded by the name $t_c$ of their core. \textsc{Name} and \textsc{Verify} are the VLM's naming loop and its yes-or-no question (\cref{sec:supp_vlm}). When the VLM confirms none of a proposal's posters under its name, the core is empty and the proposal gives no detector. Finally, \textsc{Merge} on detectors also merges two of them when half of their first $k$ posters, from the most central one, coincide, for some $k$ from $\MergeKMin$ to $\NTypical$. A merged group becomes one core and grows again.

\myparagraph{Features}\label{sec:supp_features}
As shown in \cref{fig:data} (top), we first detect the text of each poster and inpaint it. The inpainted poster is then resized to $\FeatSide\times\FeatSide$ pixels and passed once through DINOv3 ViT-L/16~\cite{simeoni2025dinov3}, to extract the $1{,}024$-d class token of the whole poster, and a $\FeatGrid\times\FeatGrid$ grid of patch tokens. We define the $\WinPerPoster$ windows of a poster as squares of this grid: nine of $\WinSmall\times\WinSmall$ patches on a $3\times3$ layout and four of $\WinLarge\times\WinLarge$ patches on a $2\times2$ layout, with a stride of $\WinStride$ patches, so that neighboring windows overlap by half their side. To describe a window, we project its patch tokens on a fixed PCA basis ($1{,}024 \to \PcaDim$), average them over $\PoolGrid\times\PoolGrid$ cells, and concatenate the averages into a $2{,}048$-d vector. The PCA basis is fit once, on $\PcaFitPatches$ unit patch tokens of $\PcaFitPosters$ posters drawn at random from the corpus. The intuition of why we don't rely on t-sne for our method is given on the bottom left part of \cref{fig:data} where the windows of one of our motif detectors ``people inside a car'' with consistent visual appearance, are close in feature space but spread over the t-sne map. 

\myparagraph{Compute}
We run our method on one machine with one NVIDIA RTX PRO 6000 Blackwell GPU (96 GB), which also serves the VLM. On \CineName, the recorded stages of our run, from the DBSCAN over windows and whole posters to the final merge, including the SVMs and the VLM's answers during growth, take $\HoursOursPlain$ hours. This does not count the
text removal, the DINOv3 features and the naming of the proposals, each computed once beforehand. On \DiscName, the growth takes $\GpuDaysDiscGrow$ GPU-days on H100 GPUs, out of which $\GpuDaysDiscGrowVlm$ correspond to the VLM's answers and
$\GpuDaysDiscGrowRest$ to the SVMs and the scoring of the covers, both of which can be parallelized.

\myparagraph{The vision--language model}\label{sec:supp_vlm}
We use Qwen3-VL-8B-Instruct~\cite{bai2025qwen3vl}, served with vLLM~\cite{kwon2023vllm} with greedy decoding, and limit a caption to $\CapTokens$ tokens, a name to $\NameTokens$ and an answer to $\CheckTokens$. The model always sees one whole poster, with its text inpainted and its longest side at most $\VlmSide$ pixels. We never crop a window from it nor draw a box on it. \cref{tab:prompts} gives our four prompts, and \cref{fig:vlm_examples} follows one proposal through them. Each of the up to $\NName$ most central posters of a proposal is captioned alone. Then, the model names what the captions share and checks the name on each of these posters. When the name fails on some of them, the model is asked again and is told which ones fail the name, until a name holds on all of them or the same name comes back. In that case, we keep the name that holds on the most of these posters (the earliest on a tie).  Every poster of the proposal is then checked against the kept name: those it fails on are left out of the core, and the detector never asks about them again. An answer is limited to $\CheckTokens$ tokens, and we count it as yes if it begins with ``yes'' (\eg, ``Yes'' or ``Yes.''), and as a no otherwise. A name is limited to $\NameTokens$ tokens; we keep its first line, in lower case, cut to $\NameWordsCut$ words.

\myparagraph{Size of the bigger motif}
\cref{fig:niche_vote} repeats \cref{fig:niche} with another horizontal axis, where each motif's size is counted as the size of the biggest motif its posters belong to. This reencoding of the x-axis, while not being directly what a motif detector shows, clarifies more a clear niche to cliche axis, which is validated with the relocation of the plot's examples.

\section{\CineName}
\label{sec:supp_data}\label{sec:supp_gt}
Besides its film's title and release year ($\MetaYear$ posters), each poster is also associated to its genres ($\MetaGenre$), country of origin ($\MetaCountry$) and director(s) ($\MetaDirector$) metadata from TMDB. \cref{fig:dist_posters} shows how the posters are distributed over time with median being around 2007.

\begin{figure}[h!]
\centering
\includegraphics[width=\linewidth]{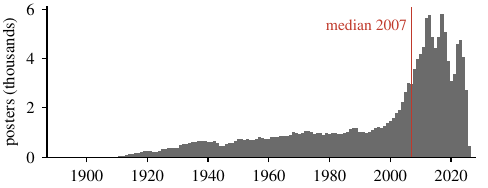}\\[1pt]
\includegraphics[width=\linewidth]{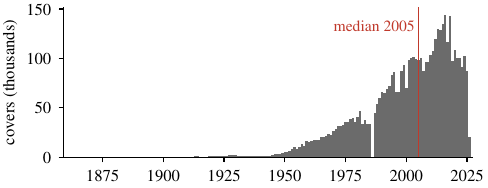}
\caption{\textbf{The datasets by year.} Number of posters of \CineName (top) and of covers of \DiscName (bottom) with a known release year, per year (red: the median). The gap at 1986 is a bug from the Discogs website, which holds a single release of that year.}
\label{fig:dist_posters}\label{fig:dist_covers}
\end{figure}

\myparagraph{Annotation}
To build our annotated subset, we start from the Letterboxd lists~\cite{gomezrosado_lists}, which gather films whose posters share a device such as hands, eyes or legs. Matching their films to our posters by title and release year gives $\GTListPosters$ posters in $\GTCategories$ categories. Since a category usually covers several motifs (a clenched fist and a hand print are both hands), we split the categories by hand into subtropes, each corresponding to one motif. We then extended categories and subtropes with $\GTAddedTool$ posters that an annotation tool that proposed  nearest neighbors of their posters, each accepted or rejected by hand, and with $\GTAddedPages$ posters found while inspecting the clusters of an early experiment. We also removed $\GTRemoved$ posters of the lists. This results in $\NGTPosters$ annotated posters.  We keep as subtropes the $\NSubtropes$ with at least five posters (\cref{tab:subtropes}), which hold $\NSubtropePosters$ posters, $\NSubtropeFromLists$ of which come from the lists.

\begin{table}[h!]
\centering
\footnotesize
\setlength{\tabcolsep}{3.5pt}
\begin{tabular}{@{}lp{2.5cm}rrrr@{}}
\toprule
Trope & \leavevmode Subtrope & Posters & Singh-like & Rooy & Ours \\
\midrule
\SubtropeRows
\bottomrule
\end{tabular}
\caption{\textbf{Subtropes.} For each label, we see the number of annotated posters and how many of them the cluster each method pairs with it as in \cref{tab:ap} (-- denotes no pairing).}
\label{tab:subtropes}
\end{table}

\begin{figure*}[t]\crplace{p}
\centering
\includegraphics[width=1.0\linewidth]{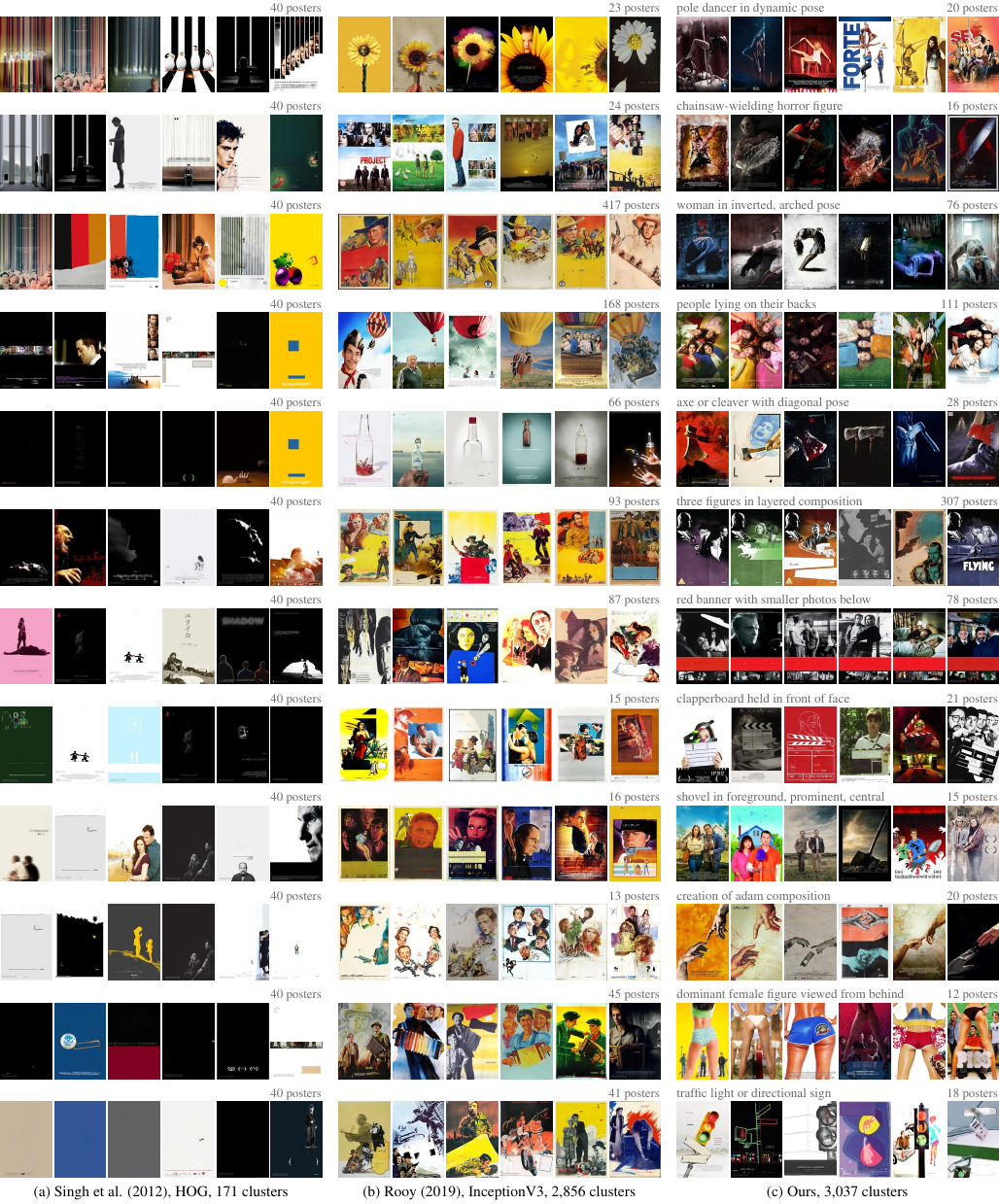}
\crfill{10.19pt}\caption{\textbf{Baselines with their original features.}  This figure extends \cref{fig:baselines} with the procedure of \Singh\ on HOG and \Rooy\ on InceptionV3, next to ours on DINOv3; for every method, distinctiveness is measured by probe SVMs on
DINOv3.}
\label{fig:baselines_orig}
\end{figure*}

\begin{table}[t]
\centering
\begin{minipage}[c]{\columnwidth}
\centering\small\setlength{\tabcolsep}{3pt}
\begin{tabular}{@{\hspace{\tabcolsep}}llrrrr@{}}
\toprule
 & & & Posters & \multicolumn{2}{c}{AP (motifs), \cref{tab:ap}} \\
\cmidrule(l){5-6}
Clustering & Growth & Detectors & (\%) & Letterboxd & Courtois \\
\midrule
\AblationRows
\bottomrule
\end{tabular}
\caption{\textbf{Ablation.} All components of different methods ablated as combinations on \CineName and tested on Letterboxd and on Courtois by mean AP (bold best across column) and, in parentheses, the number of paired motifs, as in \cref{tab:ap}. Without dimensionality reduction; t-SNE and UMAP in
\cref{tab:embed_tsne,tab:embed_umap}. Shaded: the settings of \protect\shadebox{singhbg}{Singh-like} and of
\protect\shadebox{oursbg}{ours}.}
\label{tab:ablation}
\end{minipage}
\end{table}%
\myparagraph{Mean average precision}
To compute the mean average precision of \cref{tab:ap}, each cluster ranks all the annotated posters of a source by its score: ours and the Singh-like detectors use their own SVMs, and for each of Rooy's clusters we fit an SVM on its posters as for ours (\cref{sec:growth}). The average precision of a motif for a cluster is then the mean,
over the motif's posters, of the precision at their ranks. Each motif is paired with one cluster holding at least $\kmin$ of its posters, and each cluster with at most one motif, greedily by average precision; a motif left without a cluster scores $0$. The table reports the mean over the motifs and, in parentheses, the number of paired motifs. As a sanity check, random clusters of our detectors' sizes, one detector per annotated poster, and a single detector for all posters score at most $\APSanityMax$.

\myparagraph{Found} A method finds a motif when the average precision above pairs the motif with one of the method's clusters, which happens when a cluster holds at least $\kmin$ of the motif's posters and is not taken by another motif. \Cref{tab:subtropes}, the ablation tables and \cref{fig:baselines_stats}~(c) count found motifs this way. The main text also gives a stricter count, where a cluster finds a motif only if at least $\kmin$ of the motif's posters make up at least half of the cluster's labeled posters. Only the posters of the Letterboxd lists are labeled, and the lists do not include every poster that shows a motif, so both rules see only part of what a cluster holds. The two rules ask different questions. The pairing asks whether a method has a cluster that retrieves the motif, whatever else that cluster holds. A cluster that mixes several motifs can thus still be paired with one of them, and its lower average precision then shows the mix. The stricter rule asks whether the motif is what the cluster is about. It ignores ranking and rejects mixed clusters, so that a catch-all cluster finds no motif, but neither does a mixed cluster that holds many posters of one, such as Rooy's cluster with $\RooyFaceCropHeld$ of the $\RooyFaceCropN$ face crop posters among its $\RooyFaceCropLabeled$ labeled ones. Every motif found by the stricter rule is thus also paired, and the stricter count is never higher. Ours pair $\PairedOursPlain$ motifs and find $\FoundOursPlain$ by the stricter rule (\PairedNotFoundOursPlain{} is only paired), Rooy's $\PairedRooy$ and $\FoundRooy$ (\PairedNotFoundRooy{}), and the Singh-like detectors $\PairedSingh$ by both. Among the $\FoundOursPlain$ motifs ours find by the stricter rule, whole-poster detectors find $\FoundWhole$ and window detectors $\FoundWin$, \FoundOnlyWinList{} only the latter. Over all $\PairFoundSettings$ settings of the ablation, the stricter count is lower in $\PairFoundDiffer$, by $\PairFoundGapMedian$ motifs in the median and $\PairFoundGapMax$ at most.

\section{True baselines}
\label{sec:supp_baselines} \myparagraph{\Rooy} Rooy embeds the InceptionV3~\cite{szegedy2016rethinking} features of film posters in two dimensions with t-SNE~\cite{vandermaaten2008tsne} and clusters the resulting map with DBSCAN~\cite{ester1996density}.  We follow these steps on our inpainted posters, with his features (InceptionV3 with ImageNet weights, its pooled $2{,}048$-d output at $\InceptionSide\times\InceptionSide$ pixels) and, for the main paper, with DINOv3 class tokens: we use exact t-SNE with the default settings of openTSNE~\cite{policar2024opentsne}, then DBSCAN with $\RooyMin$ points per neighborhood and a radius equal to the median distance to the $\RooyMin$th nearest neighbor on the map. This gives $\ClustersRooyIncep$ clusters with InceptionV3 vs. $\ClustersRooy$ with DINOv3.
\begin{table}[t]
\centering
\centering\small\setlength{\tabcolsep}{2pt}\renewcommand{\arraystretch}{0.86}
\begin{tabular}[t]{@{\hspace{\tabcolsep}}cllrrrr@{}}
\toprule
 & & & & Posters & \multicolumn{2}{c}{AP (motifs), \cref{tab:ap}} \\
\cmidrule(l){6-7}
$d$ & Clustering & Growth & Detectors & (\%) & Letterboxd & Courtois \\
\midrule
\EmbedRowsTsne
\bottomrule
\end{tabular}
\caption{\textbf{Dimensionality reduction (t-SNE).} As \cref{tab:ablation}, with the clustering run on a t-SNE map of the posters in $d$ dimensions. Every map is exact; at $d = 2$ it is \protect\shadebox{rooybg}{Rooy}'s, as in \cref{tab:ap}. Each source's mean AP and, in
parentheses, its number of paired motifs. }
\label{tab:embed_tsne}
\end{table}
\begin{table}[t]
\centering
\centering\small\setlength{\tabcolsep}{2pt}\renewcommand{\arraystretch}{0.86}
\begin{tabular}[t]{@{\hspace{\tabcolsep}}cllrrrr@{}}
\toprule
 & & & & Posters & \multicolumn{2}{c}{AP (motifs), \cref{tab:ap}} \\
\cmidrule(l){6-7}
$d$ & Clustering & Growth & Detectors & (\%) & Letterboxd & Courtois \\
\midrule
\EmbedRowsUmap
\bottomrule
\end{tabular}
\caption{\textbf{Dimensionality reduction (UMAP).} As in \cref{tab:embed_tsne}.}
\label{tab:embed_umap}
\end{table}
\begin{figure*}[t]
\centering
\includegraphics[width=1.0\linewidth]{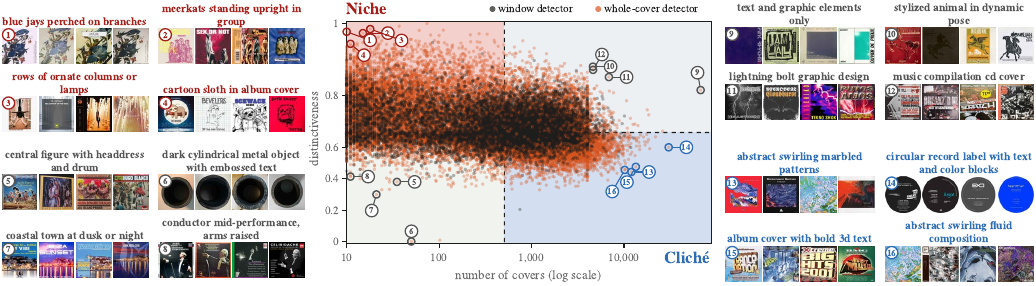}
\caption{\textbf{Album covers: niche vs.\ cliché.}  This is the equivalent of \cref{fig:niche} for the \DiscName album covers: each detector with at least $\DiscMinSize$ covers is a dot, placed according to its number of covers and its distinctiveness, and four detectors from each corner are numbered, each shown with its most central covers and, for a window detector, the window its SVM selects, boxed. Series are excluded and covers from different labels and artists. The column of dots at $\GrowCap$ covers comes from the detectors that reached the growth cap of $\GrowCap$ covers, plus their number of core proposals (smaller being $\kmin=4$).}
\label{fig:discs_niche}
\vspace{4pt}%
\end{figure*}
\begin{figure*}[t]
\centering
\begin{minipage}[t]{0.47\linewidth}\centering
\includegraphics[width=1.0\linewidth]{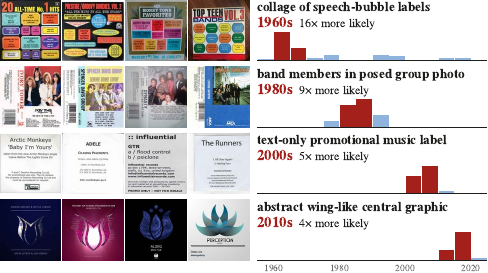}
\caption{\textbf{Album covers: detectors of a period.}  As \cref{fig:period}, for the album covers and without series: for four decades, the detector whose covers were most often made in that decade, with its most central covers from the decade (from different labels and artists) and the release years of all its covers (red: the decade).}
\label{fig:figd2}
\end{minipage}\hfill
\begin{minipage}[t]{0.47\linewidth}\centering
\includegraphics[width=1.0\linewidth]{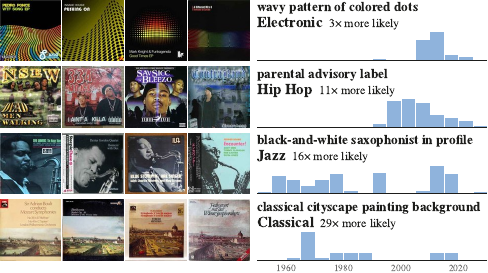}
\caption{\textbf{Album covers: detectors of a genre.}  As \cref{fig:genre}, for the album covers: for four Discogs genres, the detector whose covers most often have that genre, with its most central covers of the genre and the release years of all its covers.} 
\label{fig:figd3}
\end{minipage}
\vspace{6pt}%
\end{figure*}%

\myparagraph{\Singh} Our goal here is to show something as close as possible to the original method, so we change as little components as we can. The original samples HOG~\cite{dalal2005histograms} patches at several scales, about $150$ per image, clusters them with $k$-means, $k$ being a quarter of the patches, and refines each cluster over a few rounds, in which an SVM trained on one half of the data, against natural-world images, re-forms the cluster from its top firings on the other half, then the detectors are  ranked by purity and discriminativeness, and the top ones are shown. At its scale the method is infeasible on our corpus: one half of our posters gives about \SinghFullPatches{} patches and $k$ about \SinghFullK{}, that is, as many SVMs to train and fire in each round: scaling the time of version (b) below by the number of clusters, these rounds alone would take at least $\SinghFullDays$ days on our GPU, against $\DaysOursPlain$ days for our method. We therefore run two versions, which differ in their features and scale. (a)~\emph{Singh et al.\ on HOG}, the true baseline (\cref{fig:baselines_orig}a): their features and rounds on our $\WinPerPoster$ windows per poster, $\SinghTrueRounds$ rounds, each re-forming a cluster from its top $\SinghTrueTop$ firings, after $k$-means with $K = \SinghHogK$ over $\SinghTrueSample$ sampled windows, with negatives from our corpus; keeping every ranked detector with its $\SinghTrueTopN$ best posters gives $\ClustersSinghHogTrue$ detectors.  (b)~\emph{Singh-like}, the ablation of the main paper: the same steps on our DINOv3 windows and on every window of one half: $k$-means with $k = 4\sqrt{S} = \SinghK$ over all $S = \SinghS$ windows ($\SinghIters$ iterations, to convergence), then $\SinghRounds$ rounds in which each cluster's SVM, trained against our random windows with one round of hard negatives, fires on the other half and keeps its $\SinghTop$ best firings above $-1$; the clusters then grow up to their SVM's decision boundary, without the VLM.  Version (a) keeps the original features, version (b) our representation, so that against ours only the mining differs. Run on HOG with all the windows of (b), the method collapses: $\SinghHogDet$ detectors on $\SinghHogPosters$ posters, which find none of the ground truth motifs. Like the original paper and code, neither version merges detectors.

\begin{figure*}[t]
\centering
\includegraphics[width=1.0\linewidth]{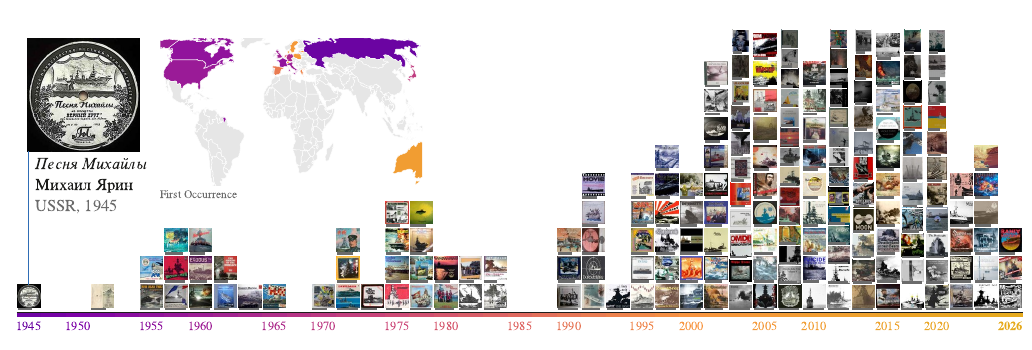}
\caption{\textbf{Album covers: the warship through time.} As \cref{fig:time}, for the album covers: the $\DiscTimeN$ covers of the ``warship depicted in maritime scene'' detector arranged by release year, each column covering two years, gray bar shows confidence $\hat{s}_d(x)$ (see \cref{eq:z}) with highest predictions being at the bottom; a column with more covers than fit piles them up in depth. The first cover is enlarged with its title, artist, country and year, and the map colors each country by the year of its first cover in the detector. The first reference is from the USSR.}
\label{fig:discs_time}
\end{figure*}%
\begin{figure}[t]
\centering
\includegraphics[width=1.0\linewidth]{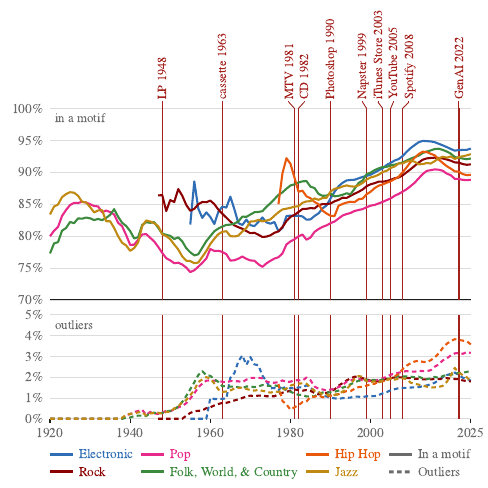}
\caption{\textbf{Album covers: motifs and outliers by genre through time.} As \cref{fig:outliers_curve}, for the six most common genres in Discogs: the yearly share of covers held by a visual motif detector (top, solid) and of outliers (bottom, dashed), each on its own axis, as they differ by more than an order of magnitude on the covers. Vertical lines mark important events of music production and distribution.}
\label{fig:discs_trends}
\end{figure}%
\myparagraph{Ablation}
To really gauge what components of each method matter we run a combination of dimensionality reduction, clustering, growth, in \cref{tab:ablation,tab:embed_tsne,tab:embed_umap} on \CineName, all on DINOv3 features. As dimensionality reduction, we compare none (\cref{tab:ablation}) to t-SNE (\cref{tab:embed_tsne}) and UMAP~\cite{mcinnes2018umap} (\cref{tab:embed_umap}), in $d = 2$, $5$, $15$ or $50$ dimensions. We add to our clustering methods the recent $e$-means~\cite{bessa2026emeans}, a replacement of Lloyd's $k$-means that
updates the centroids from mini-batches with exponential moving averages, designed to reach a lower $k$-means loss in about a tenth of the passes. We run it for five passes, with the same $k$ as $k$-means, in its spherical form on the unit DINOv3 features and in its Euclidean form on t-SNE and UMAP maps, and as growth methods we compare to the one used before that is Singh-like, version (b) above. No clustering method wins all other methods in both sets nor in discovered detectors. The best combinations grow with the VLM: $e$-means without reduction ranks Letterboxd's motifs best, and DBSCAN on the 50-dimensional t-SNE map ranks Courtois's motifs best, while on the 2-dimensional map it finds the most of them. Our method is close to the best on Letterboxd but, on Courtois, behind DBSCAN on the t-SNE maps grown with the VLM. Detectors that cover every poster are not a goal in themselves: without growth, $k$-means and $e$-means on the t-SNE and UMAP maps put nearly every poster in some detector, whether it shares a motif with the others or not, and still rank Letterboxd's motifs below ours and Courtois's no better. Our method strikes a balance: its detectors hold $\PostersPctOurs\%$ of the posters, between the $\PostersPctDbscan\%$ of DBSCAN alone and all of them. The growth matters most: for every clustering without reduction, our VLM-validated growth finds at least as many motifs as no growth or Singh's rounds, and ranks both sources at least as well. It also improves the proposals of the other methods, from Rooy's t-SNE clusters to $k$-means and $e$-means. On the t-SNE maps and on UMAP's it does so in most combinations, the exceptions being mostly Letterboxd's AP against Singh's rounds (DBSCAN's on the 2-dimensional map among them), $k$-means against no growth on Letterboxd on the 2-dimensional map, and DBSCAN on UMAP. Singh's rounds help less: without reduction they improve DBSCAN and $k$-means on both sets, but on every t-SNE map they lower the AP on Courtois, for every clustering, and on UMAP they raise DBSCAN's AP on Letterboxd but mostly lower the AP on Courtois. Without growth, t-SNE helps every clustering method: on every map, its detectors pair at least as many motifs of both sources as those of the same method without reduction, and rank them better. UMAP is generally worse than t-SNE: at the same dimension, its detectors rank both sources lower in most combinations, and none of them reaches the best AP of t-SNE on either source. Counting both sources, UMAP's detectors pair fewer motifs than those of t-SNE at the same dimension in most combinations, and in every one with DBSCAN, and no dimension is the best throughout, for either embedding. With the VLM's growth, t-SNE raises the AP on Courtois for every clustering and every map, but lowers it on Letterboxd for $e$-means and, on most maps, for $k$-means.
\section{DISCOGS-4.7M}
\label{sec:supp_discs}
\myparagraph{Corpus} Starting from $\NDiscAll$ disc covers, we remove near-duplicates (a whole-cover DINOv3 cosine of at least $\DiscDesignBar$) over all pairs of covers, which leaves $\NDiscCovers$ covers. We chose this threshold from $\DiscJudged$ pairs of covers judged by being the most consistently reliable in set of randomly picked examples. Each cover comes with its Discogs release year ($\DiscMetaYear$ covers), genres ($\DiscMetaGenre$), styles ($\DiscMetaStyle$), country ($\DiscMetaCountry$), artists ($\DiscMetaArtist$) and labels ($\DiscMetaLabel$), making a total of $\NDiscCovers$ discs (\cref{fig:dist_covers}).

\myparagraph{Run}
We apply our method (\cref{sec:method}) on disc covers with three differences. First, the windows use a coarser descriptor (nine windows of $16\times16$ patches and four of $22\times22$, with a stride of $6$, each pooled over a $2\times2$ grid into $512$ dimensions), so that the $\NDiscWindows$ windows fit on one GPU, as $512$-d half precision of a total of $\DiscWindowsGB$~GB.
Second, no text inpainting is applied to the covers. Third, a detector grows by at most one cover per work, artist and label. Growth adds at most $\DiscCap$ covers to a detector, and $\NDiscAtCap$ detectors hold at least that many. From $\NDiscSeeds$ proposals, our method gives $\NDiscDet$ detectors ($\NDiscDetWin$ window and $\NDiscDetWhole$ whole-cover ones) holding $\NDiscDetCovers$ covers (median $\DiscSizeMedian$), using $\NDiscAnswers$ answers of the model.

\myparagraph{Analysis}
\cref{fig:discs_niche} extends \cref{fig:niche} to the album covers, placing the $\NDiscRanked$ detectors of at least
$\DiscMinSize$ covers by size and distinctiveness. Niche covers show blue, meerkats standing upright, or rows of ornate columns or lamps, while clichés include album covers with bold 3D text, or abstract swirling fluid compositions. Lightning bolts and animals in dynamic poses are distinctive yet common.  Similarly to \cref{fig:period,fig:genre}, \cref{fig:figd2,fig:figd3} show the detectors typical of four decades, from collages of speech-bubble labels in the 1960s to wing-like central graphics in the 2010s, and of four genres, such as a parental advisory label for hip hop or a black-and-white saxophonist for jazz. As said before these figures leave out series, i.e., detectors whose
ten typical covers mostly come from one label or artist or are near-copies of one artwork.
\cref{fig:discs_time} follows one motif through time, as \cref{fig:time}. We first detect warships on a Soviet record of $\DiscTimeYearA$, followed by \DiscTimeOrderShort, and they become a common motif after 2000 ($\DiscTimeSince$ of their covers), spanning $\DiscTimeCountries$ countries, half of the covers from the US and Germany, and mostly rock ($\DiscTimeRock$), from punk to black metal. \cref{fig:discs_trends} repeats \cref{fig:outliers_curve} for the six most common Discogs genres. Unlike the posters, the covers show no early peak. The share held by a motif is lowest in the 1950s for pop, folk and jazz, in the 1970s for rock and electronic and in $\DiscTrendHipHopMin$ for hip hop, and grows since in every genre, to $\DiscTrendLateLo$--$\DiscTrendLateHi\%$ from 2005 on, while outliers stay at most $\DiscTrendOutMax\%$, the most in hip hop and pop since 2015. Early poster motifs are concentrated in one country, more than random posters of their time ($\CtryConcEarly$ vs $\CtryConcRandEarly$ before 1940), but cover motifs are not. A cover motif's top country holds a median $\DiscCtryEarly$ of ten of its covers before 1970 and $\DiscCtryLate$ from 2005 on, almost as much as what it holds for ten random covers of the same years ($\DiscCtryRandEarly$ and $\DiscCtryRandLate$).

\newpage
\fi
\FloatBarrier
\end{document}